\pdfoutput=1

\documentclass[11pt]{article}

\usepackage[final]{acl}
\usepackage{graphicx}

\usepackage{times}
\usepackage{latexsym}

\usepackage{amssymb}

\usepackage[T1]{fontenc}

\usepackage[utf8]{inputenc}

\usepackage{microtype}

\usepackage{inconsolata}

\usepackage[normalem]{ulem}
\usepackage{algorithm}
\usepackage{multirow}
\usepackage{amsmath} 
\usepackage{enumitem}

\usepackage{amsmath}
\usepackage{booktabs}

\usepackage{fontawesome5}

\usepackage{xcolor}

\usepackage{algorithm}
\usepackage{algpseudocode}
\usepackage[most]{tcolorbox}

\newcommand{\tk}[1]{<|#1|>\allowbreak{}}
\usepackage{cuted}  

\newtcolorbox{promptbox}[1]{
    colback=gray!8,          
    colframe=gray!60!black,  
    colbacktitle=gray!60!black, 
    coltitle=white,          
    fonttitle=\normalsize\rmfamily,
    title={#1},             
    sharp corners,
    rounded corners=all,
    arc=4pt,                 
    boxrule=1pt,             
    left=6pt, right=6pt, top=6pt, bottom=6pt,
    breakable                
}

\title{Retrievable Gradients: Continual Post-Training Without \\ Cumulative Weight Drift}

\usepackage{authblk}
\author[1,2]{\textbf{Weihang Su}\thanks{Email: oneal2000@126.com}}
\author[1]{\textbf{Jiacheng Kang}}
\author[1]{\textbf{Jingyan Xu}}
\author[2,1]{\textbf{Qingyao Ai}\thanks{Corresponding Author: aiqy@tsinghua.edu.cn}}
\author[1]{\textbf{Jianming Long}}
\author[1]{\\ \textbf{Hanwen Zhang}}
\author[1]{\textbf{Bangde Du}}
\author[1]{\textbf{Xinyuan Cao}}
\author[1]{\textbf{Min Zhang}}
\author[1]{\textbf{Yiqun Liu}}

\affil[1]{Department of Computer Science and Technology, Tsinghua University}
\affil[2]{Quancheng Laboratory}

\begin{document}
\maketitle

\begin{abstract}
Continual post-training enables models to absorb emerging knowledge after deployment, but repeatedly updating shared parameters can accumulate weight drift, potentially causing catastrophic forgetting and degrading general capabilities. Retrieval-augmented generation avoids such parameter drift, yet often lacks the depth of parametric knowledge integration. In this paper, we propose \textsc{ReGrad} (\underline{Re}trievable \underline{Grad}ients), a new paradigm that treats gradients as retrievable units of knowledge. 
\textsc{ReGrad} pre-computes document-specific gradients offline, stores them in an indexed Gradient Bank, and retrieves only query-relevant gradients at inference time for temporary weight adaptation. 
However, raw language-modeling gradients are optimized for token-level document reconstruction rather than for query-driven knowledge use. 
We therefore introduce a bi-level meta-learning objective that reshapes document-derived gradients into generalizable adaptation signals for downstream tasks.
Experiments across general and domain-specific settings show that \textsc{ReGrad} outperforms CPT and RAG baselines, enabling scalable and reversible parametric knowledge injection without accumulating weight drift\footnote{We have open-sourced all the code, data, and models at: \url{https://github.com/oneal2000/ReGrad}}.
\end{abstract}

\section{Introduction}
\label{sec-intro}

A fundamental challenge in deploying large language models (LLMs) is the mismatch between their static parameters and the dynamic nature of real-world knowledge~\citep{lazaridou2021mind,petroni2019language}.
As new information emerges after deployment, models require mechanisms that can update their knowledge without sacrificing their foundational capabilities~\citep{jang2021towards}.
Continual post-training (CPT) has emerged as a prevalent approach for this purpose, as it enables models to internalize new knowledge directly into their parametric memory~\citep{gururangan2020don}.
However, this strength is tied to an inherent limitation: CPT incorporates new knowledge by repeatedly updating the same shared parameters.
As updates accumulate over time or across large-scale corpora, the model weights gradually drift away from their original parameter state, which can lead to catastrophic forgetting and degradation of general capabilities such as reasoning and instruction following~\citep{xia2024less,lu2025fine,lin2024mitigating,cheng2023adapting,jindal2024balancing,wang2025learning}.
Moreover, CPT accumulates document-induced parameter updates into a single globally shared parameter state.
However, each downstream query typically requires only a small subset of the updated knowledge. Since the resulting parameters are shared across all queries, the knowledge relevant to a given query is inevitably mixed with many unrelated parameter modifications.
Together, these limitations expose a central tension in knowledge updating: how can we obtain the benefits of parametric knowledge integration without continuously accumulating weight drift?

Retrieval-augmented generation (RAG)~\citep{lewis2020retrieval} offers an alternative way to incorporate external knowledge by retrieving relevant textual knowledge at inference time rather than modifying model parameters.
This design naturally avoids parameter drift and helps preserve the base model's general capabilities.
Nevertheless, RAG leaves new knowledge outside the model's parameters and relies on in-context knowledge injection, where external knowledge is provided as part of the input context rather than incorporated into the model's parameters~\citep{su2025parametric}.
This distinction is important because in-context and parametric knowledge integration operate through different computational mechanisms.
In-context knowledge injection introduces new information only through the input context, where it influences the model through input-dependent inference-time computations rather than persistent parameter updates. In contrast, prior studies suggest that much of an LLM's world knowledge and reasoning capabilities are encoded in its parameters~\citep{yu-ananiadou-2024-neuron,nanda2023fact}.
Consequently, externally injected knowledge may not be utilized in the same way as knowledge internalized in the model's parameters~\citep{su2025parametric,tan2025dynamic}.
Thus, CPT and RAG present complementary strengths but also a clear trade-off: CPT provides parametric integration but accumulates irreversible updates, whereas RAG is reversible and selective but remains non-parametric.

In this work, we revisit the basic unit of parametric knowledge injection: the gradient.
In standard training, gradients are treated as transient optimization signals: they are computed from training instances, applied to update the model weights, and then discarded~\citep{rumelhart1986learning}.
Yet a gradient is more than an optimization signal: it also encodes how the model parameters should change in response to a particular document or training instance.
This observation suggests a different view: rather than permanently accumulating the effects of all document-induced gradients in a single globally shared parameter state, we can store these gradients offline as reusable parametric adaptation units, retrieve a query-relevant subset at inference time, and use them to induce query-specific parameter adaptation.
Such a mechanism retains the parametric nature of CPT by incorporating new information through parameter-space adaptation, while preventing updates from being persistently accumulated across unrelated documents or queries.

Based on this idea, we propose \textsc{ReGrad} (\underline{Re}trievable \underline{Grad}ients), a new paradigm that treats gradients as retrievable units of knowledge. 
\textsc{ReGrad} first pre-computes document-specific gradients offline and stores them in an indexed Gradient Bank. 
At inference time, given a user query, \textsc{ReGrad} retrieves the most relevant gradients and temporarily applies them to the model weights for query-specific adaptation. 
After generation, the adapted weights are discarded, and the base model is restored. 
In this way, \textsc{ReGrad} decouples knowledge acquisition from permanent weight modification: knowledge can be injected into the model parameters when needed, while the shared base parameters remain unchanged across queries.
Importantly, \textsc{ReGrad} can be used either as a standalone parameter-only adaptation method, where no retrieved text is placed in the input context, or together with in-context RAG, where retrieved gradients are applied to the model while retrieved documents are provided as textual evidence.

A remaining challenge is that raw gradients from standard language-modeling objectives are not necessarily suitable for downstream knowledge use. 
Next-token prediction encourages document reconstruction, whereas downstream tasks require query answering, implicit fact extraction, and reasoning over document content. 
As a result, simply caching language-modeling gradients may overfit surface-level token patterns rather than improve query-driven generalization. 
To address this problem, \textsc{ReGrad} introduces a bi-level meta-learning objective that reshapes document-derived gradients into generalizable adaptation signals. 
The inner update applies the document-induced gradient to the model, while the outer objective evaluates the adapted model on downstream supervision. 
Through this objective, \textsc{ReGrad} meta-learns gradient formation and application, so that temporary adaptation improves downstream behavior rather than merely reconstructing document text.

We conduct extensive experiments across both general and domain-specific settings, including law and medicine. 
In the parameter-only setting, \textsc{ReGrad} outperforms continual post-training and other parametric updating baselines, demonstrating that document-induced gradients can serve as effective, retrievable adaptation units.
When combined with in-context knowledge injection, \textsc{ReGrad} further improves performance, suggesting that retrievable gradients and textual contexts provide complementary forms of knowledge injection.
These results position \textsc{ReGrad} not as a replacement for RAG, but as a reversible parameter-side mechanism that can be used independently or integrated with existing retrieval-augmented systems.
In summary, this paper makes the following contributions:

\begin{itemize}[leftmargin=*, nosep]

\item We propose \textsc{ReGrad}, a new paradigm that stores gradients as retrievable units, enabling query-specific parametric updating without permanently modifying the model parameters.

\item We introduce a bi-level meta-learning objective that reshapes document-derived language-modeling gradients for knowledge use rather than document reconstruction.

\item We empirically show that \textsc{ReGrad} outperforms CPT and RAG and can be combined with in-context RAG to achieve complementary gains.

\end{itemize}

\begin{figure*}[t]
\centering
\includegraphics[width=\textwidth]{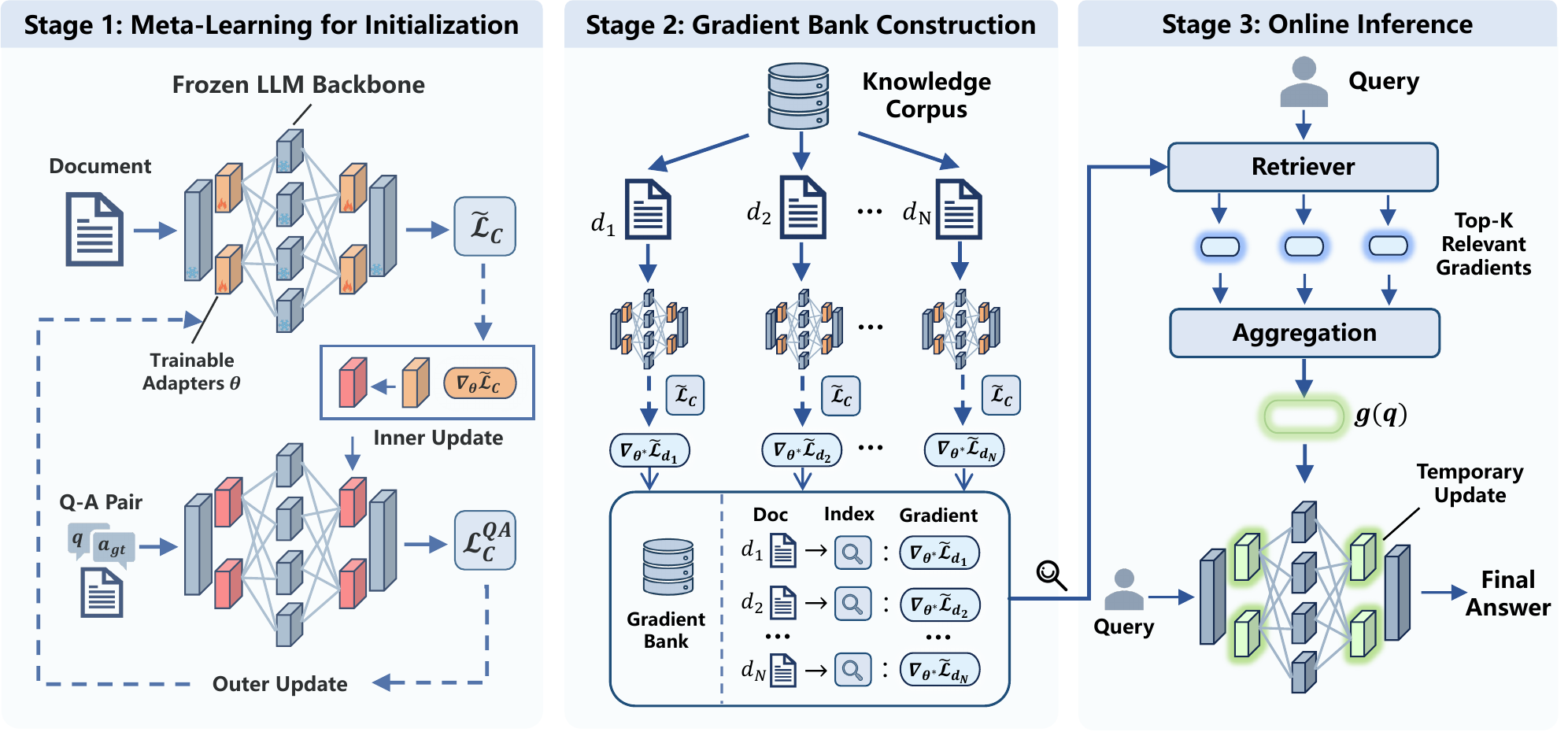}
\caption{
{Overview of the \textsc{ReGrad} framework.}
The method operates in three stages: Meta-Learning, Gradient Bank Construction, and Online Inference.
}
\label{fig:pipeline}
\end{figure*}

\section{Related Work}
\label{sec:related}

\textsc{ReGrad} is related to three lines of work: continual post-training (CPT), retrieval-augmented generation (RAG), and test-time training. 
CPT injects new knowledge into model parameters, but repeated updates may cause catastrophic forgetting and capability drift, motivating mitigation strategies such as regularization, distillation, and replay~\citep{kirkpatrick2017overcoming,li2017learning,lopez2017gradient}. 
Knowledge editing also modifies parameters, but aims for localized factual updates with limited side effects~\citep{meng2022locating,meng2022mass,mitchell2021fast,mitchell2022memory}. 
Unlike these methods, \textsc{ReGrad} retrieves temporary, query-relevant parameter shifts without permanently changing the base model. 
RAG instead conditions generation on retrieved evidence~\citep{lewis2020retrieval,guu2020retrieval,borgeaud2022improving}. 
The closest line is Parametric RAG, which converts documents into retrievable parameter modules~\citep{su2025parametric}. 
Finally, test-time training performs online gradient updates during inference~\citep {hardttest,chen2023reckoning}, whereas \textsc{ReGrad} avoids online backpropagation by retrieving precomputed gradients. 
Further review and discussion of related work is provided in Appendix~\ref{app:related}.

\section{The ReGrad Framework}
\label{sec:method}

In this section, we introduce our proposed \textsc{ReGrad} framework, shown in \autoref{fig:pipeline}.
We first present the overview of our entire framework in \S~\ref{sec:overview}.
We then describe our proposed framework, which operates in a three-stage lifecycle:
(1) {Meta-learning} to align unsupervised document gradients with downstream QA tasks (\S ~\ref{sec:stage1});
(2) {Gradient Bank Construction} to store these aligned gradients offline (\S ~\ref{sec:stage2}); and
(3) {Online Inference} to perform retrieval-augmented parametric updates (\S ~\ref{sec:stage3}).
Unless otherwise specified, \textsc{ReGrad} denotes the parametric-only setting without retrieved passages in the QA prompt.

\subsection{Overview}
\label{sec:overview}

Let $\mathcal{M}_W$ denote a pre-trained LLM with base parameters $W$, and let $\mathcal{D}=\{d_i\}_{i=1}^N$ be an external knowledge corpus.
Given a test-time query $q$, \textsc{ReGrad} aims to inject relevant knowledge from $\mathcal{D}$ through a temporary parametric update, while restoring the model to $\mathcal{M}_W$ after generation.
The framework consists of three stages.
First, a meta-learning stage learns adaptation-friendly parameters that turn document-induced gradients into QA-oriented updates.
Second, these parameters are used to build a Gradient Bank by pre-computing document-specific gradients.
Third, at inference time, query-relevant gradients are retrieved and aggregated to produce a temporary update in the parameter space.

\subsection{Stage 1: Meta-learning for Initialization}
\label{sec:stage1}

In this stage, \textsc{ReGrad} follows a MAML-style bi-level optimization~\cite{finn2017model}: an \emph{inner} unsupervised update extracts knowledge from retrieved documents, and an \emph{outer} QA loss evaluates whether the updated model can answer downstream questions without seeing the documents in the prompt.
The goal of Stage~1 is to learn an initialization and step sizes such that
{an unsupervised language-modeling gradient computed from a document}
can be used as a {generalizable} one-step adaptation for downstream question answering.
This stage is the key to making stored gradients usable beyond surface-form reconstruction: during deployment, we will only have access to the knowledge corpus documents (no QA labels), yet we want the retrieved gradient updates to consistently improve query-answering behavior.
To make document-specific gradients storable and retrievable at the corpus scale, \textsc{ReGrad} avoids computing or storing gradients over the full model parameters $W$.
Instead, we introduce a compact adaptation subspace by inserting LoRA adapters into selected transformer blocks and restricting adaptation to their parameters $\theta$.
The resulting model is $\mathcal{M}_{(W,\theta)}$, where $W$ remains frozen, and all document-induced gradients are computed with respect to $\theta$.
This low-rank parameterization substantially reduces the storage cost of the Gradient Bank while retaining sufficient capacity to encode instance-level knowledge updates.

\subsubsection{Inner objective.}
Given a document (or chunk\footnote{For our main experiment, we treat each $d_i$ as a chunked passage of length up to 200 tokens.})
$C=\{c_t\}_{t=1}^{T}$, we compute its gradient representation by differentiating the standard language modeling objective:

{\small
\begin{equation}
\label{eq:lm_loss}
\tilde{\mathcal{L}}_C(\theta)
= -\frac{1}{T}\sum_{t=1}^{T}\log p_{(W,\theta)}(c_t \mid c_{<t}).
\end{equation}}

\noindent We denote the resulting gradient in the plug-in space as a {knowledge atom}, and apply a virtual update step:

{\small
\begin{equation}
\label{eq:inner_update}
g_C = \nabla_{\theta}\tilde{\mathcal{L}}_C(\theta),
\qquad
\theta_C = \theta - \alpha \odot g_C ,
\end{equation}
}

\noindent where $\alpha$ is a set of meta-learned \emph{scalar} step sizes (one scalar per trainable LoRA tensor), and $\odot$ broadcasts each scalar to scale the corresponding gradient tensor.
Importantly, $\tilde{\mathcal{L}}_C$ is QA label-free and depends only on $C$,
which later allows us to compute the gradient for {every} document in the corpus.
However, a raw gradient from Eq.~\eqref{eq:lm_loss} mainly optimizes surface-form reconstruction, since next-token prediction rewards verbatim continuation~\cite{allen2023physics, su2025parametric}.
Stage~1 therefore meta-learns $\theta$ and $\alpha$ so that the same label-free gradient step becomes useful for downstream QA rather than merely reconstructing the document text.

\subsubsection{Outer Objective.}
While the inner loop operates on individual documents unsupervised, the outer loop evaluates whether these updates collectively enable the model to answer queries.
We assume a meta-training set consisting of tuples $(\mathcal{C}, q, a)$. Here, $\mathcal{C} = \{C_k\}_{k=1}^K$ is a set of context passages containing the necessary facts, and $(q, a)$ is a QA pair grounded in $\mathcal{C}$.
In a retrieval scenario, a query typically retrieves multiple documents. To simulate this during meta-training and ensure the learned $\alpha$ and $\theta$ can handle multi-source knowledge, we aggregate the gradients from multiple retrieved passages.
Specifically, we compute the \emph{knowledge atom} $g_{C_k}$ for each passage individually (as in Eq.~\ref{eq:inner_update}) and sum them to form a unified update direction:

{\small
\begin{equation}
\label{eq:aggregation}
g_{\mathcal{C}} = \sum_{k=1}^{K} g_{C_k},
\qquad
\theta_{\mathcal{C}} = \theta - \alpha \odot g_{\mathcal{C}} .
\end{equation}
}

\noindent A critical design constraint in \textsc{ReGrad} is that during the outer pass, the context $\mathcal{C}$ is {excluded} from the model's input prompt. 
The model must generate the answer $a$ strictly from the query $q$, relying solely on the parametric knowledge injected into $\theta_{\mathcal{C}}$.
This forces the meta-optimization to treat $\theta$ as a "gradient shaper": it must learn to encode factual content into the gradient $g_{\mathcal{C}_q}$ rather than relying on in-context knowledge injection.
The outer loss is thus the standard conditional likelihood of the answer:

{\small
\begin{equation}
\label{eq:outer_loss}
\mathcal{L}^{\text{QA}}_{\mathcal{C}}(\theta_{\mathcal{C}})
= -\frac{1}{L}\sum_{t=1}^{L}\log p_{(W,\theta_{\mathcal{C}})}(a_t \mid q, a_{<t}).
\end{equation}
}

\subsubsection{Meta-learning Objective.}
Finally, we formally define the meta-learning objective. We optimize the initial parameters $\theta$ and step sizes $\alpha$ to minimize the outer QA loss after applying the aggregated unsupervised update. The complete bi-level optimization problem is:

{\small
\begin{equation}
\label{eq:meta_objective}
\min_{\theta,\alpha}\;
\mathbb{E}_{(\mathcal{C},q,a)\sim \mathcal{D}_{\text{meta}}}
\left[
\mathcal{L}^{\text{QA}}_{\mathcal{C}}
\Big(\theta - \alpha \odot \sum_{C\in\mathcal{C}}g_{C}(\theta)\Big)
\right].
\end{equation}
}

\noindent Eq.~\eqref{eq:meta_objective} differentiates through the inner update in Eq.~\eqref{eq:inner_update}, so the meta-gradient propagates through the dependence of the document gradient $g_C(\theta)=\nabla_{\theta}\tilde{\mathcal{L}}_C(\theta)$ on the initialization $\theta$.
Consequently, downstream QA supervision does not directly optimize the model to answer questions; instead, it shapes how unsupervised language-modeling gradients are computed.
From an optimization perspective, meta-training encourages each unsupervised document gradient to function as a task-relevant one-step update: although $g_C$ is computed without QA labels, its direction is learned to reduce the QA loss after aggregation and application.
In this sense, $\theta$ acts as a \emph{gradient shaper}, transforming language-modeling gradients into adaptation directions that are useful for question answering.
Note that once meta-training is complete, $(\theta^*, \alpha^*)$ are fixed and used to generate the gradient bank in Stage~2.

\subsection{Stage 2: Gradient Bank Construction}
\label{sec:stage2}

Given the meta-learned initialization and step sizes $(\theta^*, \alpha^*)$, Stage~2 constructs a \emph{Gradient Bank} that converts the entire knowledge corpus into a collection of retrievable gradients.
For each document $d_i \in \mathcal{D}$, we compute an unsupervised language-modeling gradient in the plug-in parameter space using the meta-learned initialization: $ g_i = \nabla_{\theta}\tilde{\mathcal{L}}_{d_i}(\theta^*).$
This gradient $g_i$ serves as a compact \emph{parametric knowledge representation} of the factual knowledge in $d_i$.
Crucially, due to the meta-learning stage (Stage~1), these gradients are reshaped to better support downstream question answering rather than merely optimizing surface-form reconstruction.
As a result, each $g_i$ can be interpreted as a \emph{knowledge atom} that captures how the model parameters should change in order to internalize the information in $d_i$.

To enable query-time retrieval, we associate each gradient $g_i$ with its corresponding document text $d_i$ and build a retrieval index over the corpus. The choice of indexing strategy is orthogonal to the ReGrad framework and depends on the downstream retrieval setup: lexical methods (e.g., inverted indices with BM25) index raw document text, while dense retrievers index learned document embeddings. At inference time, retrieval operates solely over document representations, and the retrieved document IDs are then used to fetch their stored gradients.

\subsection{Stage 3: Online Inference}
\label{sec:stage3}
At inference time, given a user query $q$, we retrieve top-$K$ most relevant documents $\{d_{i_k}\}_{k=1}^{K}$ and fetch their stored knowledge atoms $\{g_{i_k}\}_{k=1}^{K}$ from the Gradient Bank.
We combine retrieved knowledge atoms by {plain gradient accumulation}:

{\small
\begin{equation}
\label{eq:agg}
g(q) = \sum_{k=1}^{K} g_{i_k},
\qquad
\theta_q = \theta^* - \alpha^* \odot g(q).
\end{equation}
}

\noindent The answer is generated using $\mathcal{M}_{(W,\theta_q)}$ without appending retrieved passages,
and after generation, we revert to $\theta^*$, ensuring that no updates accumulate across queries.

The aggregation mechanism in Eq.~\eqref{eq:agg} follows directly from the additivity of gradients.
In standard training, techniques such as Distributed Data Parallelism and Gradient Accumulation combine gradients computed from different samples or micro-batches to obtain the gradient of the corresponding summed or averaged loss.
Similarly, since all stored knowledge atoms are computed at the same meta-learned initialization $\theta^*$, summing the gradients of the retrieved documents is equivalent to computing the gradient of their aggregate language-modeling objective at $\theta^*$.
Therefore, accumulating the retrieved gradients is equivalent to taking the gradient of the aggregate language-modeling objective with respect to the retrieved documents.
The resulting update follows the same additive structure as standard gradient-based training, while the meta-learned step sizes control how the aggregated gradient is applied.

\begin{table*}[t]
\centering

\resizebox{\textwidth}{!}{
\begin{tabular}{llcccccccccc}
\toprule

 &  &  \multicolumn{3}{c}{\textbf{General}}& \multicolumn{3}{c}{\textbf{Law}}&\multicolumn{3}{c}{\textbf{Medical}} & \multicolumn{1}{c}{\textbf{Overall}} \\
\cmidrule(lr){3-5} \cmidrule(lr){6-8} \cmidrule(lr){9-11} \cmidrule(lr){12-12}
\textbf{Model} & \textbf{Method} &  \textbf{\textsc{2WQA}} &  \textbf{\textsc{CWQ}} &\textbf{\textsc{HQA}} & \textbf{\textsc{CHOLD}} & \textbf{\textsc{LHF}} & \textbf{\textsc{Housing}} &\textbf{\textsc{PubMed}} & \textbf{\textsc{MedQA}} & \textbf{\textsc{BioASQ}} & \textbf{Avg.} \\
\midrule
\multirow{10}{*}{\textbf{LLaMA-1B}}
& \textbf{Direct} & 21.27*& 31.71*& 16.08*& 42.1*& 41.1*& 64.3*& 17.3*& 6.24*& 42.6*& 31.41*\\
& \textbf{Standard CPT} & 15.34*& 28.19*& 11.42*& 40.1*& 40.3*& 59.7*& 18.9*& 6.13*& 46.6*& 29.63*\\
& \textbf{Instruction CPT} & 30.71*& 34.74*& 18.93*& 40.7*& 47.1*& 62.9*& 20.1*& 6.80*& 41.4*& 33.71*\\
& \textbf{RAG (ICL)} & 24.08*& 33.58*& 24.01*& 42.7*& 41.2*& 65.8*& 70.2*& 7.88*& 71.2*& 42.29*\\
& \textbf{PE-RAG} & 24.98*& 35.63*& 27.94*& 44.0*& 43.2*& 62.7*& 62.9*& 6.12*& 66.3*& 41.53*\\
& \textbf{PRAG} & 24.73*& 32.27*& 20.11*& 41.8*& 41.2*& 63.5*& 17.3*& 6.92*& 47.3*& 32.79*\\
& \textbf{Fine-tuned-Direct} & 27.96*& 42.74*& 18.66*& 45.2*& 86.1*& 79.2*& 79.0*& 10.63 & 72.7*& 51.35*\\
& \textbf{Fine-tuned-RAG} & 32.04*& 46.82*& \underline{31.09}*& 42.1*& 90.9*& 68.4*& 78.2*& \underline{10.82} & \underline{80.8} & 53.46*\\
& \textbf{Fine-tuned-PRAG} & 31.43*& 45.26*& 24.73*& 65.1*& 89.5*& 81.9 & 92.4*& 10.78 & 76.8*& 57.54*\\
& \textbf{ReGrad} & \underline{34.61}*& \underline{47.18}*& 30.72*& \underline{65.7} & \textbf{94.8} & \underline{82.1} & \underline{93.3} & 10.12 & 74.6*& \underline{59.24}*\\
& \textbf{ReGrad + ICL} & \textbf{40.63} & \textbf{50.34} & \textbf{39.18} & \textbf{68.4} & \underline{94.7} & \textbf{84.5} & \textbf{97.9} & \textbf{11.50} & \textbf{83.4} & \textbf{63.39} \\
\midrule
\multirow{10}{*}{\textbf{LLaMA-3B}}
& \textbf{Direct} & 16.58*& 36.69*& 18.77*& 55.8*& 46.1*& 47.7*& 88.1*& 10.63*& 77.8*& 44.24*\\
& \textbf{Standard CPT} & 10.96*& 33.94*& 15.48*& 54.2*& 44.5*& 45.7*& 78.4*& 9.47*& 75.1*& 40.86*\\
& \textbf{Instruction CPT} & 30.28*& 43.26*& 24.09*& 55.5*& 48.2*& 45.4*& 82.6*& 13.06*& 78.3*& 46.74*\\
& \textbf{RAG (ICL)} & 20.13*& 29.84*& 24.11*& 63.2*& 42.2*& 51.2*& 82.8*& 7.13*& 74.2*& 43.87*\\
& \textbf{PE-RAG} & 17.75*& 40.05*& 32.92*& 66.2*& 47.5*& 53.6*& 81.7*& 11.58*& 77.9*& 47.69*\\
& \textbf{PRAG} & 19.34*& 38.22*& 20.73*& 56.0*& 43.8*& 48.1*& 89.1*& 8.18*& 79.4*& 44.77*\\
& \textbf{Fine-tuned-Direct} & 29.07*& 47.27*& 22.88*& 45.0*& 87.7*& 79.0*& 79.1*& 15.08*& 76.1*& 53.47*\\
& \textbf{Fine-tuned-RAG} & 35.48*& 51.11 & 32.49*& 50.9*& 91.9 & 70.5*& 86.4*& 15.19*& \underline{82.1}*& 57.34*\\
& \textbf{Fine-tuned-PRAG} & 32.39*& 51.42 & 29.51*& 64.2*& 89.2*& 82.5 & 92.7 & 16.12 & 79.1*& 59.68*\\
& \textbf{ReGrad} & \underline{36.82}*& \underline{52.57} & \underline{33.43}*& \underline{68.8} & \textbf{94.7} & \underline{83.1} & \underline{92.8} & \underline{16.66} & 78.6*& \underline{61.94}*\\
& \textbf{ReGrad + ICL} & \textbf{44.18} & \textbf{54.08} & \textbf{43.47} & \textbf{71.3} & \underline{94.6} & \textbf{84.8} & \textbf{97.9} & \textbf{18.21} & \textbf{85.7} & \textbf{66.03} \\
\midrule
\multirow{10}{*}{\textbf{LLaMA-8B}}
& \textbf{Direct} & 33.42*& 42.73*& 26.34*& 61.3*& 54.1*& 46.0*& 91.2*& 16.46*& 81.2*& 50.31*\\
& \textbf{Standard CPT} & 25.67*& 40.24*& 23.51*& 58.1*& 53.8*& 46.2*& 90.3*& 15.59*& 80.2*& 48.18*\\
& \textbf{Instruction CPT} & 36.31*& 46.58*& 27.44*& 63.8*& 54.0*& 49.5*& 38.7*& 19.24*& 60.6*& 44.02*\\
& \textbf{RAG (ICL)} & 26.98*& 31.47*& 35.23*& 66.3*& 39.9*& 53.5*& 80.1*& 17.88*& 80.5*& 47.98*\\
& \textbf{PE-RAG} & 29.61*& 36.58*& 38.70*& 69.7*& 51.4*& 60.2*& 80.6*& 14.15*& 81.0*& 51.33*\\
& \textbf{PRAG} & 31.63*& 40.96*& 27.18*& 60.3*& 52.0*& 46.2*& 92.4*& 15.93*& 82.1*& 49.86*\\
& \textbf{Fine-tuned-Direct} & 32.18*& 49.67*& 28.84*& 56.9*& 84.2*& 80.2*& 92.8*& \underline{22.89} & 81.4*& 58.79*\\
& \textbf{Fine-tuned-RAG} & 37.17*& 54.21 & \underline{39.16}*& 59.6*& 91.9 & 69.5*& 92.7*& \textbf{23.22} & \underline{84.6} & 61.34*\\
& \textbf{Fine-tuned-PRAG} & 35.71*& 51.26*& 32.97*& 66.2*& 78.6*& 83.0 & 91.8*& 21.37 & 82.6*& 60.39*\\
& \textbf{ReGrad} & \underline{38.43}*& \underline{56.41} & 36.31*& \underline{70.3}*& \textbf{96.2} & \underline{84.8} & \underline{93.2}*& 21.65 & 81.9*& \underline{64.36}*\\
& \textbf{ReGrad + ICL} & \textbf{45.06} & \textbf{57.52} & \textbf{45.27} & \textbf{74.1} & \underline{94.8} & \textbf{85.4} & \textbf{98.1} & 19.83*& \textbf{85.3} & \textbf{67.26} \\
\bottomrule
\end{tabular}
}
\caption{Main results under the Top-$1000$ setting on general and domain-specific benchmarks. We compare \textbf{ReGrad} with direct generation, CPT variants, RAG, PRAG, and their fine-tuned variants across LLaMA-3.2 1B, 3B, and LLaMA-3.1 8B. The best result in each column and model block is highlighted in \textbf{bold}, and the second best is \underline{underlined}. $^\ast$ denotes a result significantly lower than the bolded result under a $t$-test ($p<0.05$). Overall Avg. is the macro average over dataset-level scores, where each dataset uses its task-specific metric.}
\vspace{-3mm}
\label{tab:main_results}
\end{table*}

\section{Experimental Setup}

\textbf{Baselines.}
We compare \textsc{ReGrad} with direct generation (no external knowledge provided), parametric knowledge injection (Standard CPT and its variant Instruction CPT~\cite{cheng2023adapting}), non-parametric knowledge injection, and a retrieval-based parameter update method (Parametric RAG~\citep{su2025parametric}). 
For non-parametric knowledge injection, we include both Standard RAG~\citep{lewis2020retrieval} and a stronger prompt-engineered (PE) variant, denoted as PE-RAG. 
Standard RAG uses the widely adopted RAG prompt adopted by FLARE~\cite{jiang2023active}, DRAGIN~\cite{su2024dragin}, and Parametric RAG~\cite{su2025parametric}, while PE-RAG differs only by replacing this default prompt with a stronger RAG-specific prompt. 
We include PE-RAG to control for the effect of prompt quality in non-parametric retrieval augmentation.
Unless otherwise specified, all main experiments and ablation studies use the PE-RAG prompt, except for Standard RAG and Parametric RAG, which follow their original prompt settings to maintain consistency with prior work.
For fairness, we also include fine-tuned variants of applicable baselines.
More details on baseline implementations and settings are described in Appendix~\ref{app:baselines}.

\noindent \textbf{Implementation Details of ReGrad.} 
\textsc{ReGrad} is instantiated on LLaMA-family models across different series and scales, including LLaMA-3.2-1B/3B and LLaMA-3.1-8B. 
We use BM25 for retrieval using Elasticsearch and aggregate the top-$K$ passages with $K=3$. 
For each domain, we meta-train a single LoRA initialization and step-size configuration using the aggregated training splits of the corresponding downstream datasets, while strictly separating meta-training and evaluation data. 
In addition to the default parametric-only \textsc{ReGrad} setting, we also evaluate \textsc{ReGrad+ICL}, a separately meta-trained variant where the retrieved passages are included in both the outer-loop QA prompt during meta-training and the generation prompt during inference.
The bi-level objective for both settings is optimized for 1 epoch with a meta-batch size of $16$, a peak outer learning rate of $4\times10^{-4}$, and bfloat16 precision. Further details, including hyperparameters and prompt templates, are provided in Appendix~\ref{app:implementation}.

\noindent \textbf{Benchmarks and Corpora.} We evaluate ReGrad on both a general-domain Wikipedia setting and two vertical domains (Medicine and Law). 
For the general domain, we use the DPR Wikipedia dump~\citep{karpukhin2020dense} as the external corpus and evaluate on 2WQA~\citep{ho2020constructing}, HotpotQA~\citep{yang2018hotpotqa}, and CWQ~\citep{talmor-berant-2018-web}. 
For vertical domains, we use PubMed Abstracts (The Pile)~\citep{2021arXiv210100027G} and Pile-of-Law ~\citep{henderson2022pile} as the corpus, and evaluate on PubMedQA~\citep{jin2019pubmedqa}, MedQA~\citep{jin2021disease}, and BioASQ~\citep{tsatsaronis2015overview}, CaseHOLD~\citep{zheng2021does}, Learned Hands Family~\citep{guha2023legalbench}, and HousingQA~\citep{2025arXiv250503970Z}. More details on datasets and corpora are provided in Appendix~\ref{app:benchmarks}.

\noindent \textbf{Evaluation Protocols and Metrics.}
We use Accuracy for closed-ended tasks and token-level F1 for open-ended QA. 
All results use greedy decoding with official HuggingFace chat templates.
The main results are reported for the first 1,000 test instances of each dataset, with other experiments reported for the first 300.
More details on evaluation and metrics are provided in Appendix~\ref{app:metrics}.

\section{Experimental Results}

\subsection{Main Results}

Table~\ref{tab:main_results} compares \textsc{ReGrad} with parametric updating baselines (CPT and PRAG) and non-parametric retrieval baselines (RAG). 
We evaluate \textsc{ReGrad} in two modes: a parameter-only mode using only retrieved gradients, and a hybrid mode, \textit{\textsc{ReGrad} + ICL}, which additionally places the retrieved documents in the input context.
In the parameter-only mode, \textsc{ReGrad} achieves the best average performance among non-hybrid methods across all three model scales, showing that document-induced gradients provide an effective reversible adaptation signal. 
While \textsc{ReGrad} does not uniformly outperform the fine-tuned RAG baseline on every individual benchmark, its strong overall performance demonstrates the value of retrieved gradients alone.
The hybrid \textit{\textsc{ReGrad} + ICL} setting achieves the best average performance across all model scales and the strongest results on most benchmarks. 
These results show that retrieved gradients complement in-context evidence, positioning \textsc{ReGrad} as either a standalone parameter-only adaptation method or a parameter-side module for retrieval-augmented generation.

\begin{table}[t]
\centering

\setlength{\tabcolsep}{4pt}
\resizebox{\columnwidth}{!}{
\begin{tabular}{lcccccccc}
\toprule
\multirow{2}{*}{\# \textbf{Docs}} 
& \multicolumn{2}{c}{\textbf{2Wiki}} 
& \multicolumn{2}{c}{\textbf{CWQ}} 
& \multicolumn{2}{c}{\textbf{HotpotQA}} 
& \multicolumn{2}{c}{\textbf{Avg.}} \\
\cmidrule(lr){2-3}
\cmidrule(lr){4-5}
\cmidrule(lr){6-7}
\cmidrule(lr){8-9}
& \textbf{CPT} &\textbf{RG} 
& \textbf{CPT} & \textbf{RG} 
& \textbf{CPT} & \textbf{RG} 
& \textbf{CPT} & \textbf{RG} \\
\midrule

\textbf{2.5K} 
& 27.14 & {33.95}
& 43.32 & {55.91}
& 21.73 & {23.26}
& 30.73 & {37.71} \\

\textbf{5K}   
& 28.26 & {34.12}
& 44.55 & {57.16}
& 21.43 & {27.65}
& 31.41 & {39.64} \\

\textbf{7.5K} 
& 27.43 & {36.12}
& 45.72 & {59.84}
& 22.46 & {31.84}
& 31.87 & {42.60} \\

\textbf{10K } 
& 20.25 & {37.70}
& 45.89 & {60.10}
& 17.26 & {34.35}
& 27.80 & {44.05} \\

\bottomrule
\end{tabular}
}
\vspace{-2mm}
\caption{
Capability preservation under continual knowledge injection on LLaMA-8B.
We report F1 on three benchmarks.
RG denotes \textsc{ReGrad}. 
}
\vspace{-4mm}
\label{tab:capability}
\end{table}

\subsection{Cumulative Drift Analysis}
\label{sec:capability}

We further examine whether \textsc{ReGrad} avoids the cumulative drift observed in continual post-training as knowledge injection scales.
On LLaMA-3.1-8B, we build a 10K-document corpus by combining the top-3 retrieved documents for all downstream queries with randomly sampled noise documents, and incrementally inject the shuffled corpus into the model.
CPT uses the same LoRA architecture and all applicable hyperparameters as \textsc{ReGrad}.
The difference is that CPT continuously updates model parameters, while \textsc{ReGrad} stores document-induced gradients and applies only retrieved gradients as temporary query-specific updates.
As shown in Table~\ref{tab:capability}, CPT initially improves performance but eventually degrades as more documents are injected, suggesting the effect of cumulative parameter drift.
In contrast, \textsc{ReGrad} consistently improves across all datasets as the Gradient Bank grows, because a larger bank increases the likelihood of retrieving relevant gradients while each query still applies only a one-step, temporary update.
The average trend further confirms that \textsc{ReGrad} scales knowledge injection without the long-term degradation observed in CPT.

\begin{figure}[t]
\centering
\includegraphics[width=\columnwidth]{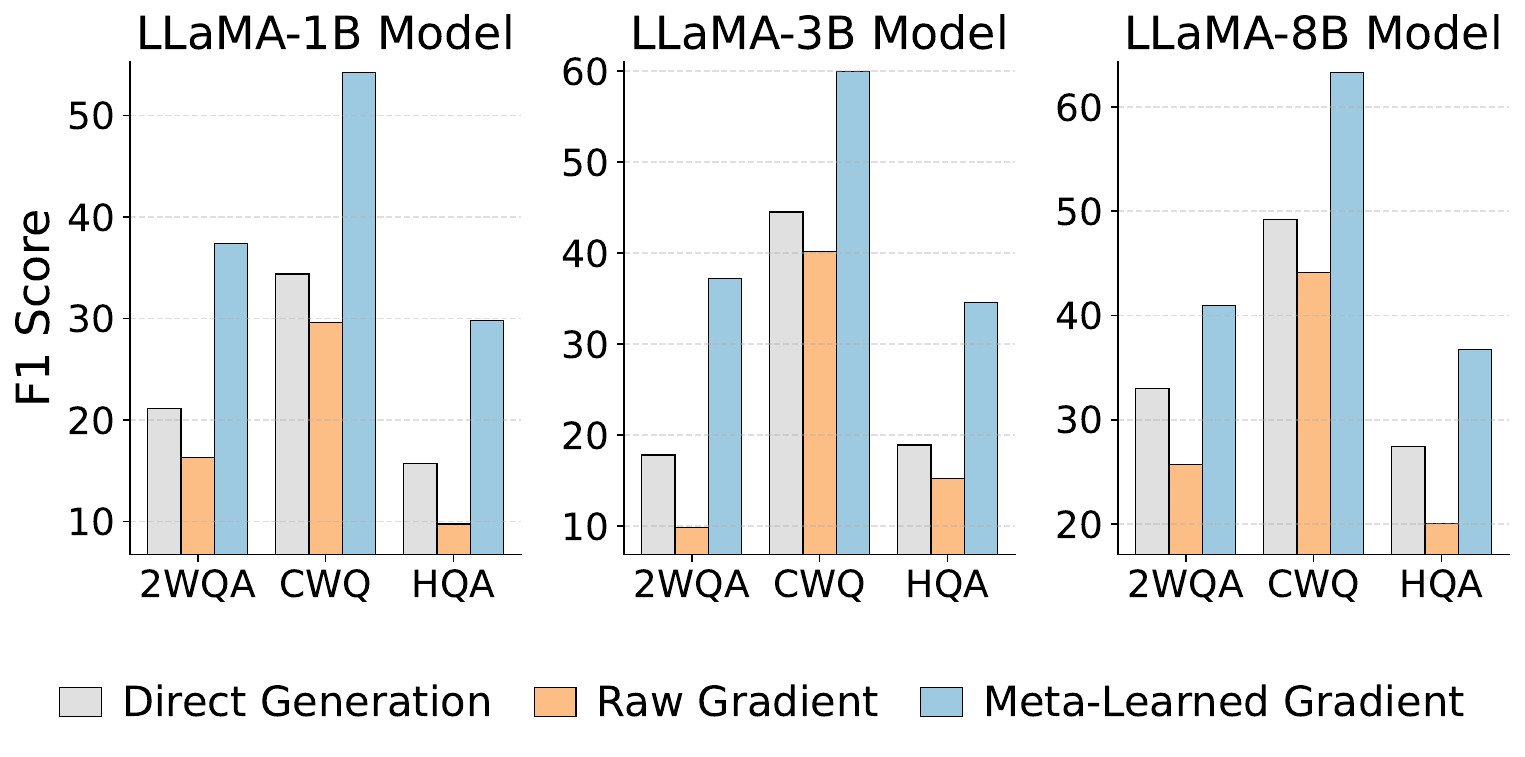}
\caption{Impact of the meta-learning stage. 
}
\label{fig:raw_gradient}
\end{figure}

\subsection{Ablation Studies}

\paragraph{Necessity of Meta-Learning for Gradient Alignment.} 
To validate the effectiveness of our bi-level optimization, we compare ReGrad against the Raw Gradient baseline, where gradients are derived directly from the standard next-token prediction objective without meta-training. 
As illustrated in \autoref{fig:raw_gradient}, the Raw Gradient approach consistently underperforms Direct Generation across all model scales and benchmarks. 
This performance degradation highlights a fundamental misalignment: raw gradients are optimized for surface-form reconstruction (verbatim memorization), which, when applied naively, introduces optimization noise that disrupts the model's pre-trained capabilities. This result confirms that effective knowledge injection cannot be achieved by simply caching optimization signals. ReGrad’s meta-learning is therefore indispensable, successfully transforming gradients from disruptive reconstruction signals into aligned adaptation artifacts that enhance downstream tasks.

\begin{figure}[t]
\centering
\includegraphics[width=\columnwidth]{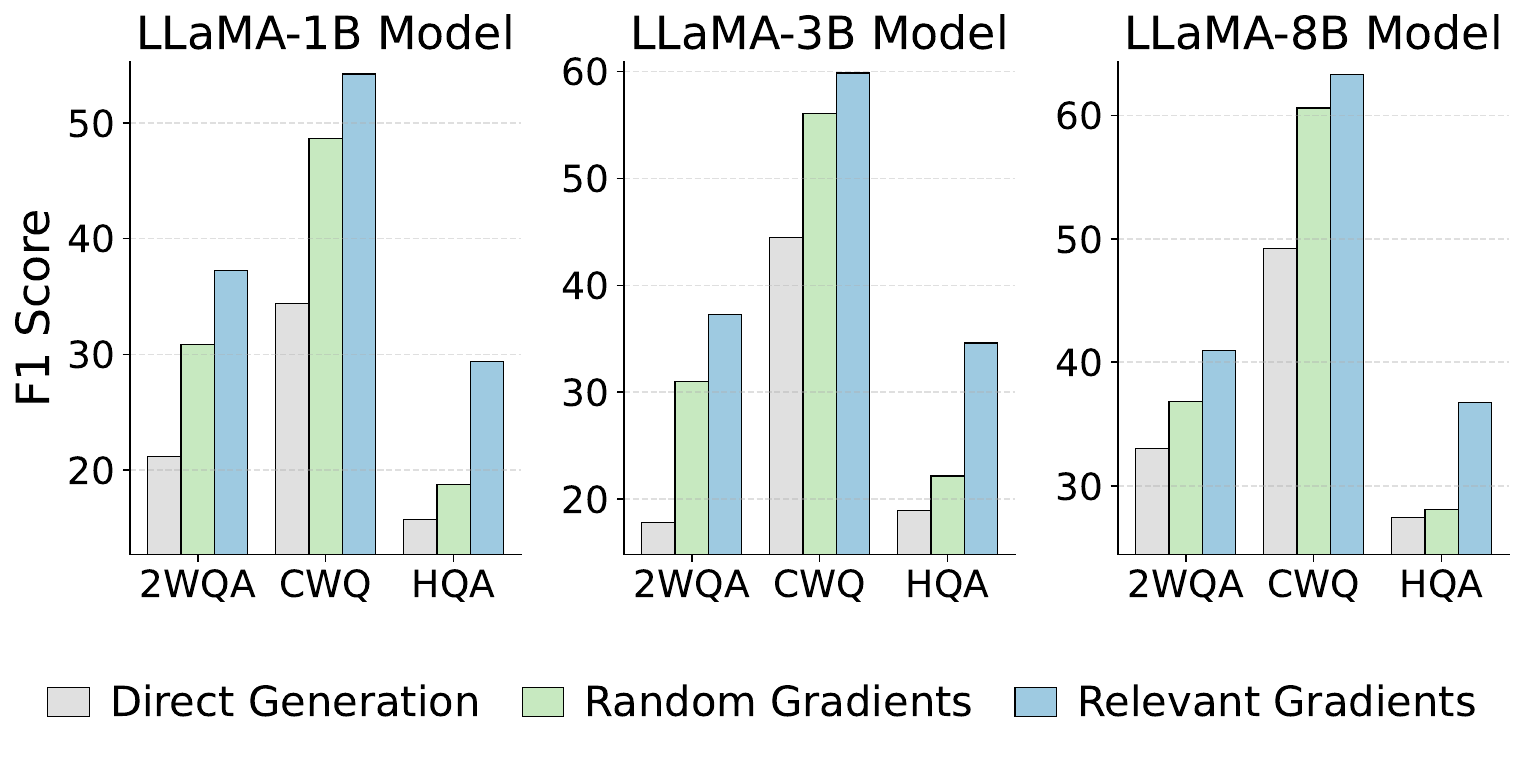}
\caption{Impact of gradient relevance. 
}
\label{fig:random}
\end{figure}

\paragraph{Impact of Gradient Relevance.}
To investigate whether the performance gains stem from specific knowledge injection or merely a meta-learned task adaptation, we conduct an ablation study by replacing the retrieved gradients with randomly sampled gradients from the Gradient Bank (referred to as "Random Gradients"). As illustrated in Figure~\ref{fig:random}, applying random gradients yields only a marginal improvement over the direct generation baseline, while injecting relevant gradients remains significantly superior.
This observation implies that the gradient signal carries both a generic task-solving schematic and specific factual knowledge. Random gradients only trigger a general "QA mode," explaining the marginal performance gain. Crucially, the significant gain from relevant gradients confirms that specific semantic information is the key performance driver, validating that stored gradients effectively encode unique document knowledge into the LLM's parametric space.

\begin{figure}[t]
\centering
\includegraphics[width=\columnwidth]{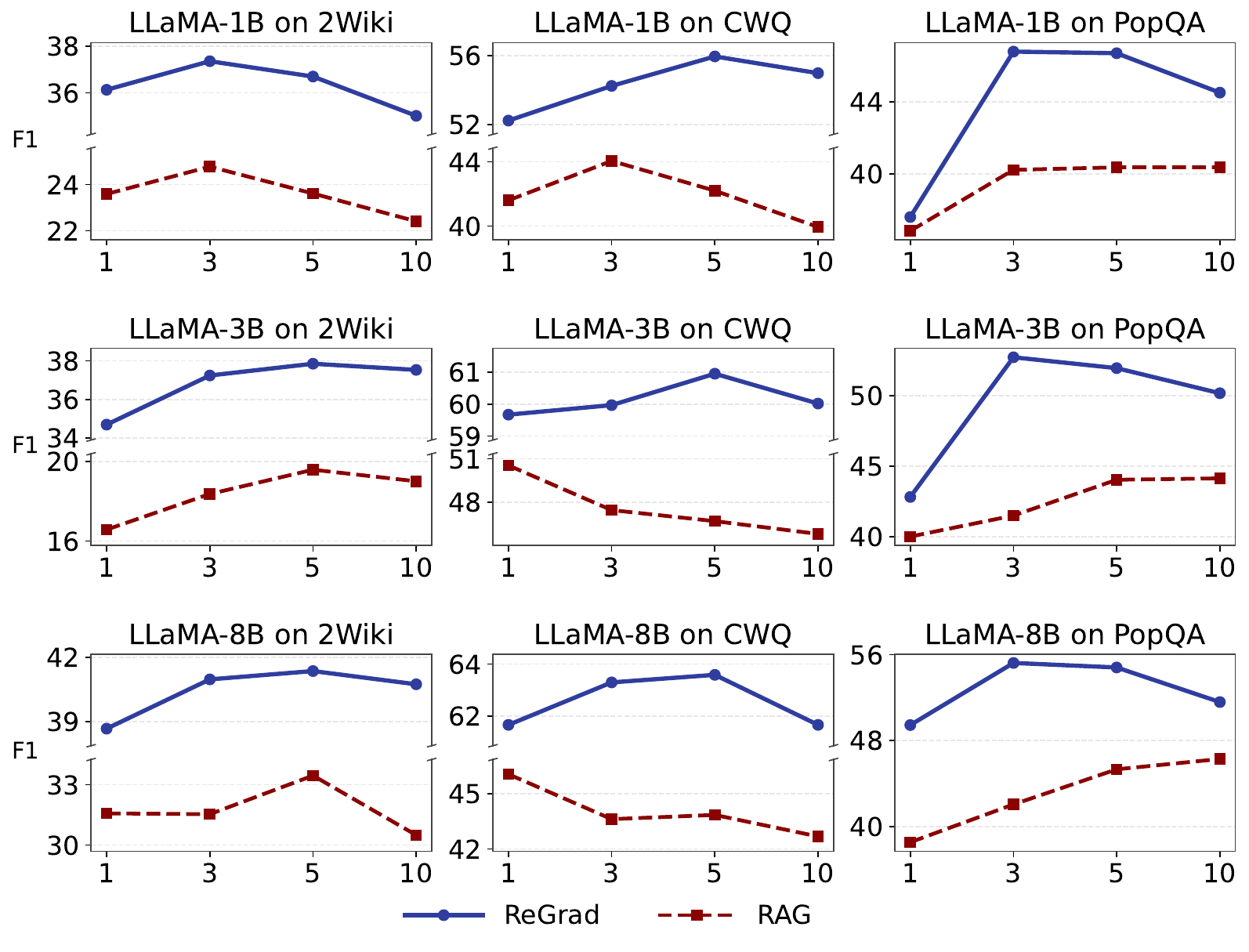}
\caption{Impact of the number of injected gradients.}
\label{fig:merge}
\end{figure}

\subsection{Effect of Injected Gradient Number}
\label{sec:merge}

We study how the number of injected gradients affects \textsc{ReGrad} by varying the number of retrieved documents $k \in \{1, 3, 5, 10\}$.
As shown in Figure~\ref{fig:merge}, \textsc{ReGrad} consistently outperforms RAG in absolute performance, while both methods exhibit an inverted-U trend.
Increasing $k$ from $1$ to $3$ improves recall and mitigates retriever failure when the gold document is not ranked first.
However, further increasing $k$ introduces different failure modes: RAG may suffer from lost-in-the-middle effects and attention dilution, whereas \textsc{ReGrad} may suffer from \textit{gradient interference}, where irrelevant gradients harm the temporary update.
The trend is also task-dependent: \textsc{ReGrad} is more robust on complex reasoning tasks such as CWQ, while RAG scales better on simple entity-centric tasks such as PopQA.
To further isolate the effect of noisy retrieval, we report an additional stress test with fixed relevant passages and random distractors in Appendix~\ref{app:noise}.

\subsection{Cross-Domain Transfer}
\label{sec:ood}

\begin{table}[t]
\centering

\resizebox{0.99\linewidth}{!}{
\begin{tabular}{llccc}
\toprule
\textbf{Model} & \textbf{Method} & \textbf{General} & \textbf{Med} & \textbf{Law} \\
\midrule
\multirow{4}{*}{\textbf{1B}} 
& \textbf{Base LLM} & 22.06 & 27.50 & 45.33 \\
& \textbf{General-init ReGrad} & 42.05 & 28.12 & 43.33 \\
& \textbf{Continued ReGrad} & 38.42 & 56.07 & 79.11 \\
& \textbf{Mixed-domain ReGrad} & 42.45 & 55.35 & 76.44 \\
\midrule
\multirow{4}{*}{\textbf{3B}} 
& \textbf{Base LLM} & 25.36 & 59.72 & 48.22 \\
& \textbf{General-init ReGrad} & 46.14 & 45.46 & 53.67 \\
& \textbf{Continued ReGrad} & 43.18 & 61.72 & 79.44 \\
& \textbf{Mixed-domain ReGrad} & 47.80 & 60.70 & 76.55 \\
\bottomrule
\end{tabular}
}
\caption{
Cross-domain transfer of \textsc{ReGrad}.
General-init applies a general-domain meta-initialization to all domains, while Continued and Mixed-domain further adapt it for domain transfer.
}\label{tab:ood_robustness}
\vspace{-3mm}
\end{table}

We further examine whether a general-domain meta-initialization can support \textsc{ReGrad} across domains.
In Table~\ref{tab:ood_robustness}, Base LLM denotes direct generation without any augmentation.
General-init \textsc{ReGrad} uses the meta-initialization learned from the general domain to construct Gradient Banks for all domains.
Although this setting performs well on the General domain, it brings limited or even negative gains on Medical and Law tasks, indicating that the meta-initialization is domain-dependent.
We then evaluate two simple transfer strategies:
\textit{Continued ReGrad}, which further adapts the general-domain initialization on target-domain data, and
\textit{Mixed-domain ReGrad}, which trains the initialization with both general and target-domain data.
Both strategies substantially improve Medical and Law performance while preserving strong General performance.
These results suggest that \textsc{ReGrad} can be transferred to specialized domains, but its meta-initialization requires adaptation.

\section{Efficiency Analysis}
\label{sec:efficiency}

We briefly discuss the efficiency of \textsc{ReGrad}; detailed discussion and complexity analysis are provided in Appendix~\ref{sec:app-efficiency}.

\setlength{\parskip}{0pt}

\paragraph{Offline computation.}
\textsc{ReGrad} shifts adaptation cost to the offline construction of the Gradient Bank.
For each document, it performs a single backward pass restricted to the low-rank plug-in parameters, making the total construction cost comparable to one epoch of CPT in the LoRA subspace.
In our measurements on an NVIDIA A100 GPU, gradient construction takes 50.97, 71.17, and 84.05 ms/doc for the 1B, 3B, and 8B models, respectively.

\paragraph{Online inference.}

Unlike RAG, which lengthens the input with retrieved documents, \textsc{ReGrad} injects knowledge through lightweight low-rank updates without adding tokens.
Applying retrieved gradients requires only tensor additions, incurring just 6-15 ms across the 1B, 3B, and 8B models.
Importantly, \textsc{ReGrad} supports batched inference despite using query-specific retrieved gradients.
This is analogous to multi-LoRA serving: requests share the same backbone while applying different LoRAs, a setting that systems such as Punica~\cite{chen2024punica} and S-LoRA~\cite{sheng2023s} have shown can be efficiently batched in practice.
Appendix~\ref{sec:app-latency} reports measured end-to-end latency across baselines and model scales.

\paragraph{Storage cost.}
The main trade-off of \textsc{ReGrad} is storage, as the Gradient Bank scales linearly with the corpus size.
In our 1B setting, each document gradient requires a few hundred KB under the low-rank plug-in parameterization (detailed analysis is in Appendix~\ref{sec:app-efficiency}).
This cost can be further reduced by caching gradients only for frequently accessed or high-value documents, while computing gradients for long-tail documents on demand.

\section{Conclusion}
In this paper, we introduced \textsc{ReGrad}, a paradigm that treats document-induced gradients as retrievable knowledge units and optimizes them through a bi-level meta-learning objective.
By retrieving and temporarily applying query-relevant gradients, \textsc{ReGrad} enables selective parametric knowledge injection without permanent weight drift.
Experiments show that \textsc{ReGrad} outperforms CPT and RAG baselines while remaining complementary to in-context RAG.

\section*{Acknowledgements}
This work is supported by the Research Project of Quancheng Laboratory, China (Grant No. QCL20250105)

\section*{Limitations}

As an initial step toward the broader paradigm of retrievable parametric adaptation, this work demonstrates the feasibility of treating gradients as retrievable knowledge units. 
Several limitations remain and point to important directions for future research.
First, the current ReGrad framework relies on task-specific supervised data during the meta-learning stage to align document-derived gradients with downstream knowledge use. Future work could explore weakly supervised or synthetic-data-based meta-training to reduce this dependence.
Second, ReGrad requires storing document-specific gradients in a Gradient Bank. Although this follows the same general assumption as Parametric RAG methods that store retrievable parametric artifacts, it introduces additional storage overhead compared with text-only retrieval. More compact gradient representations, sparsification, quantization, or selective storage strategies could further improve scalability.
Finally, our experiments focus on injecting factual knowledge for knowledge-intensive tasks. However, gradients may also encode reusable capabilities rather than only document-level knowledge. Extending ReGrad to capability injection, such as retrieving and applying gradients associated with agent skills or tool-use behaviors, is an important direction for future research.

\bibliography{custom}

@article{rumelhart1986learning,
  title={Learning representations by back-propagating errors},
  author={Rumelhart, David E and Hinton, Geoffrey E and Williams, Ronald J},
  journal={nature},
  volume={323},
  number={6088},
  pages={533--536},
  year={1986},
  publisher={Nature Publishing Group UK London}
}

@article{karpukhin2020dense,
  title={Dense passage retrieval for open-domain question answering},
  author={Karpukhin, Vladimir and O{\u{g}}uz, Barlas and Min, Sewon and Lewis, Patrick and Wu, Ledell and Edunov, Sergey and Chen, Danqi and Yih, Wen-tau},
  journal={arXiv preprint arXiv:2004.04906},
  year={2020}
}

@inproceedings{allen2023physics,
  title={Physics of Language Models: Part 3.1, Knowledge Storage and Extraction},
  author={Allen-Zhu, Zeyuan and Li, Yuanzhi},
  year={2024},
  booktitle={Forty-first International Conference on Machine Learning}
}

@article{jin2021disease,
  title={What disease does this patient have? a large-scale open domain question answering dataset from medical exams},
  author={Jin, Di and Pan, Eileen and Oufattole, Nassim and Weng, Wei-Hung and Fang, Hanyi and Szolovits, Peter},
  journal={Applied Sciences},
  volume={11},
  number={14},
  pages={6421},
  year={2021},
  publisher={MDPI}
}

@inproceedings{jin2019pubmedqa,
  title={Pubmedqa: A dataset for biomedical research question answering},
  author={Jin, Qiao and Dhingra, Bhuwan and Liu, Zhengping and Cohen, William and Lu, Xinghua},
  booktitle={Proceedings of the 2019 conference on empirical methods in natural language processing and the 9th international joint conference on natural language processing (EMNLP-IJCNLP)},
  pages={2567--2577},
  year={2019}
}

@article{wang2024lekube,
  title={LeKUBE: A Legal Knowledge Update BEnchmark},
  author={Wang, Changyue and Su, Weihang and Yiran, Hu and Ai, Qingyao and Wu, Yueyue and Luo, Cheng and Liu, Yiqun and Zhang, Min and Ma, Shaoping},
  journal={arXiv preprint arXiv:2407.14192},
  year={2024}
}

@article{ho2020constructing,
  title={Constructing a multi-hop QA dataset for comprehensive evaluation of reasoning steps},
  author={Ho, Xanh and Nguyen, Anh-Khoa Duong and Sugawara, Saku and Aizawa, Akiko},
  journal={arXiv preprint arXiv:2011.01060},
  year={2020}
}

@inproceedings{su2025dynamic,
  title={Dynamic and parametric retrieval-augmented generation},
  author={Su, Weihang and Ai, Qingyao and Zhan, Jingtao and Dong, Qian and Liu, Yiqun},
  booktitle={Proceedings of the 48th International ACM SIGIR Conference on Research and Development in Information Retrieval},
  pages={4118--4121},
  year={2025}
}

@article{tan2025dynamic,
  title={Dynamic parametric retrieval augmented generation for test-time knowledge enhancement},
  author={Tan, Yuqiao and He, Shizhu and Liao, Huanxuan and Zhao, Jun and Liu, Kang},
  journal={arXiv preprint arXiv:2503.23895},
  year={2025}
}

@inproceedings{su2025parametric,
  title={Parametric retrieval augmented generation},
  author={Su, Weihang and Tang, Yichen and Ai, Qingyao and Yan, Junxi and Wang, Changyue and Wang, Hongning and Ye, Ziyi and Zhou, Yujia and Liu, Yiqun},
  booktitle={Proceedings of the 48th International ACM SIGIR Conference on Research and Development in Information Retrieval},
  pages={1240--1250},
  year={2025}
}

@inproceedings{dong2025decoupling,
  title={Decoupling Knowledge and Context: An Efficient and Effective Retrieval Augmented Generation Framework via Cross Attention},
  author={Dong, Qian and Ai, Qingyao and Wang, Hongning and Liu, Yiding and Li, Haitao and Su, Weihang and Liu, Yiqun and Chua, Tat-Seng and Ma, Shaoping},
  booktitle={Proceedings of the ACM on Web Conference 2025},
  pages={4386--4395},
  year={2025}
}

@inproceedings{su2025judge,
  title={Judge: Benchmarking judgment document generation for chinese legal system},
  author={Su, Weihang and Yue, Baoqing and Ai, Qingyao and Hu, Yiran and Li, Jiaqi and Wang, Changyue and Zhang, Kaiyuan and Wu, Yueyue and Liu, Yiqun},
  booktitle={Proceedings of the 48th International ACM SIGIR Conference on Research and Development in Information Retrieval},
  pages={3573--3583},
  year={2025}
}

@article{wang2025decoupling,
  title={Decoupling Reasoning and Knowledge Injection for In-Context Knowledge Editing},
  author={Wang, Changyue and Su, Weihang and Ai, Qingyao and Zhou, Yujia and Liu, Yiqun},
  journal={arXiv preprint arXiv:2506.00536},
  year={2025}
}

@article{jiang2023active,
  title={Active retrieval augmented generation},
  author={Jiang, Zhengbao and Xu, Frank F and Gao, Luyu and Sun, Zhiqing and Liu, Qian and Dwivedi-Yu, Jane and Yang, Yiming and Callan, Jamie and Neubig, Graham},
  journal={arXiv preprint arXiv:2305.06983},
  year={2023}
}

@inproceedings{su2024mitigating,
  title={Mitigating entity-level hallucination in large language models},
  author={Su, Weihang and Tang, Yichen and Ai, Qingyao and Wang, Changyue and Wu, Zhijing and Liu, Yiqun},
  booktitle={Proceedings of the 2024 Annual International ACM SIGIR Conference on Research and Development in Information Retrieval in the Asia Pacific Region},
  pages={23--31},
  year={2024}
}

@inproceedings{yu-ananiadou-2024-neuron,
    title = "Neuron-Level Knowledge Attribution in Large Language Models",
    author = "Yu, Zeping  and
      Ananiadou, Sophia",
    editor = "Al-Onaizan, Yaser  and
      Bansal, Mohit  and
      Chen, Yun-Nung",
    booktitle = "Proceedings of the 2024 Conference on Empirical Methods in Natural Language Processing",
    month = nov,
    year = "2024",
    address = "Miami, Florida, USA",
    publisher = "Association for Computational Linguistics",
    url = "https://aclanthology.org/2024.emnlp-main.191/",
    doi = "10.18653/v1/2024.emnlp-main.191",
    pages = "3267--3280",
}

@online{nanda2023fact,
  author    = {Neel Nanda and Senthooran Rajamanoharan and János Kramár and Rohin Shah},
  title     = {Fact Finding: Attempting to Reverse-Engineer Factual Recall on the Neuron Level},
  year      = {2023},
  url       = {https://www.lesswrong.com/posts/iGuwZTHWb6DFY3sKB/fact-finding-attempting-to-reverse-engineer-factual-recall},
  note      = {Accessed: 2025-01-24},
}

@misc{Llama-3.2-1B-Instruct,
  title = {Llama-3.2-1B-Instruct},
  author = {Meta},
  year = {2024},
  url = {https://huggingface.co/meta-llama/Llama-3.2-1B-Instruct},
  note = {Accessed: 2024-09}
}

@article{edge2024local,
  title={From local to global: A graph rag approach to query-focused summarization},
  author={Edge, Darren and Trinh, Ha and Cheng, Newman and Bradley, Joshua and Chao, Alex and Mody, Apurva and Truitt, Steven and Larson, Jonathan},
  journal={arXiv preprint arXiv:2404.16130},
  year={2024}
}

@article{ram2023context,
  title={In-context retrieval-augmented language models},
  author={Ram, Ori and Levine, Yoav and Dalmedigos, Itay and Muhlgay, Dor and Shashua, Amnon and Leyton-Brown, Kevin and Shoham, Yoav},
  journal={arXiv preprint arXiv:2302.00083},
  year={2023}
}

@inproceedings{borgeaud2022improving,
  title={Improving language models by retrieving from trillions of tokens},
  author={Borgeaud, Sebastian and Mensch, Arthur and Hoffmann, Jordan and Cai, Trevor and Rutherford, Eliza and Millican, Katie and Van Den Driessche, George Bm and Lespiau, Jean-Baptiste and Damoc, Bogdan and Clark, Aidan and others},
  booktitle={International conference on machine learning},
  pages={2206--2240},
  year={2022},
  organization={PMLR}
}

@article{lewis2020retrieval,
  title={Retrieval-augmented generation for knowledge-intensive nlp tasks},
  author={Lewis, Patrick and Perez, Ethan and Piktus, Aleksandra and Petroni, Fabio and Karpukhin, Vladimir and Goyal, Naman and K{\"u}ttler, Heinrich and Lewis, Mike and Yih, Wen-tau and Rockt{\"a}schel, Tim and others},
  journal={Advances in Neural Information Processing Systems},
  volume={33},
  pages={9459--9474},
  year={2020}
}

@inproceedings{guu2020retrieval,
  title={Retrieval augmented language model pre-training},
  author={Guu, Kelvin and Lee, Kenton and Tung, Zora and Pasupat, Panupong and Chang, Mingwei},
  booktitle={International conference on machine learning},
  pages={3929--3938},
  year={2020},
  organization={PMLR}
}

@article{robertson2009probabilistic,
  title={The probabilistic relevance framework: BM25 and beyond},
  author={Robertson, Stephen and Zaragoza, Hugo and others},
  journal={Foundations and Trends{\textregistered} in Information Retrieval},
  volume={3},
  number={4},
  pages={333--389},
  year={2009},
  publisher={Now Publishers, Inc.}
}

@article{su2023caseformer,
  title={Caseformer: Pre-training for Legal Case Retrieval},
  author={Su, Weihang and Ai, Qingyao and Wu, Yueyue and Ma, Yixiao and Li, Haitao and Liu, Yiqun},
  journal={arXiv preprint arXiv:2311.00333},
  year={2023}
}

@article{su2024unsupervised,
  title={Unsupervised real-time hallucination detection based on the internal states of large language models},
  author={Su, Weihang and Wang, Changyue and Ai, Qingyao and Hu, Yiran and Wu, Zhijing and Zhou, Yujia and Liu, Yiqun},
  journal={arXiv preprint arXiv:2403.06448},
  year={2024}
}

@article{su2024dragin,
  title={Dragin: Dynamic retrieval augmented generation based on the real-time information needs of large language models},
  author={Su, Weihang and Tang, Yichen and Ai, Qingyao and Wu, Zhijing and Liu, Yiqun},
  journal={arXiv preprint arXiv:2403.10081},
  year={2024}
}

@article{su2024stard,
  title={STARD: A Chinese Statute Retrieval Dataset with Real Queries Issued by Non-professionals},
  author={Su, Weihang and Hu, Yiran and Xie, Anzhe and Ai, Qingyao and Que, Zibing and Zheng, Ning and Liu, Yun and Shen, Weixing and Liu, Yiqun},
  journal={arXiv preprint arXiv:2406.15313},
  year={2024}
}

@article{fang2024scaling,
  title={Scaling Laws For Dense Retrieval},
  author={Fang, Yan and Zhan, Jingtao and Ai, Qingyao and Mao, Jiaxin and Su, Weihang and Chen, Jia and Liu, Yiqun},
  journal={arXiv preprint arXiv:2403.18684},
  year={2024}
}

@inproceedings{mallen-etal-2023-trust,
    title = "When Not to Trust Language Models: Investigating Effectiveness of Parametric and Non-Parametric Memories",
    author = "Mallen, Alex  and
      Asai, Akari  and
      Zhong, Victor  and
      Das, Rajarshi  and
      Khashabi, Daniel  and
      Hajishirzi, Hannaneh",
    editor = "Rogers, Anna  and
      Boyd-Graber, Jordan  and
      Okazaki, Naoaki",
    booktitle = "Proceedings of the 61st Annual Meeting of the Association for Computational Linguistics (Volume 1: Long Papers)",
    month = jul,
    year = "2023",
    address = "Toronto, Canada",
    publisher = "Association for Computational Linguistics",
    url = "https://aclanthology.org/2023.acl-long.546",
    doi = "10.18653/v1/2023.acl-long.546",
    pages = "9802--9822",
}

@inproceedings{talmor-berant-2018-web,
    title = "The Web as a Knowledge-Base for Answering Complex Questions",
    author = "Talmor, Alon  and
      Berant, Jonathan",
    editor = "Walker, Marilyn  and
      Ji, Heng  and
      Stent, Amanda",
    booktitle = "Proceedings of the 2018 Conference of the North {A}merican Chapter of the Association for Computational Linguistics: Human Language Technologies, Volume 1 (Long Papers)",
    month = jun,
    year = "2018",
    address = "New Orleans, Louisiana",
    publisher = "Association for Computational Linguistics",
    url = "https://aclanthology.org/N18-1059",
    doi = "10.18653/v1/N18-1059",
    pages = "641--651",
   
}

@article{yang2018hotpotqa,
  title={HotpotQA: A dataset for diverse, explainable multi-hop question answering},
  author={Yang, Zhilin and Qi, Peng and Zhang, Saizheng and Bengio, Yoshua and Cohen, William W and Salakhutdinov, Ruslan and Manning, Christopher D},
  journal={arXiv preprint arXiv:1809.09600},
  year={2018}
}

@inproceedings{hulora,
  title={LoRA: Low-Rank Adaptation of Large Language Models},
  author={Hu, Edward J and Wallis, Phillip and Allen-Zhu, Zeyuan and Li, Yuanzhi and Wang, Shean and Wang, Lu and Chen, Weizhu and others},
  booktitle={International Conference on Learning Representations},
  year={2022}
}

@inproceedings{xia2024less,
  title={LESS: selecting influential data for targeted instruction tuning},
  author={Xia, Mengzhou and Malladi, Sadhika and Gururangan, Suchin and Arora, Sanjeev and Chen, Danqi},
  booktitle={Proceedings of the 41st International Conference on Machine Learning},
  pages={54104--54132},
  year={2024}
}

@article{jindal2024balancing,
  title={Balancing Continuous Pre-Training and Instruction Fine-Tuning: Optimizing Instruction-Following in LLMs},
  author={Jindal, Ishan and Badrinath, Chandana and Bharti, Pranjal and Vinay, Lakkidi and Sharma, Sachin Dev},
  journal={arXiv preprint arXiv:2410.10739},
  year={2024}
}

@article{wang2025learning,
  title={Learning Dynamics in Continual Pre-Training for Large Language Models},
  author={Wang, Xingjin and Tissue, Howe and Wang, Lu and Li, Linjing and Zeng, Daniel Dajun},
  journal={arXiv preprint arXiv:2505.07796},
  year={2025}
}

@article{lu2025fine,
  title={Fine-tuning large language models for domain adaptation: Exploration of training strategies, scaling, model merging and synergistic capabilities},
  author={Lu, Wei and Luu, Rachel K and Buehler, Markus J},
  journal={npj Computational Materials},
  volume={11},
  number={1},
  pages={84},
  year={2025},
  publisher={Nature Publishing Group UK London}
}

@article{chen2023reckoning,
  title={Reckoning: Reasoning through dynamic knowledge encoding},
  author={Chen, Zeming and Weiss, Gail and Mitchell, Eric and Celikyilmaz, Asli and Bosselut, Antoine},
  journal={Advances in Neural Information Processing Systems},
  volume={36},
  pages={62579--62600},
  year={2023}
}

@inproceedings{lin2024mitigating,
  title={Mitigating the alignment tax of rlhf},
  author={Lin, Yong and Lin, Hangyu and Xiong, Wei and Diao, Shizhe and Liu, Jianmeng and Zhang, Jipeng and Pan, Rui and Wang, Haoxiang and Hu, Wenbin and Zhang, Hanning and others},
  booktitle={Proceedings of the 2024 Conference on Empirical Methods in Natural Language Processing},
  pages={580--606},
  year={2024}
}

@inproceedings{cheng2023adapting,
  title={Adapting large language models via reading comprehension},
  author={Cheng, Daixuan and Huang, Shaohan and Wei, Furu},
  booktitle={The Twelfth International Conference on Learning Representations},
  year={2023}
}

@ARTICLE{2021arXiv210100027G,
       author = {{Gao}, Leo and {Biderman}, Stella and {Black}, Sid and {Golding}, Laurence and {Hoppe}, Travis and {Foster}, Charles and {Phang}, Jason and {He}, Horace and {Thite}, Anish and {Nabeshima}, Noa and {Presser}, Shawn and {Leahy}, Connor},
        title = "{The Pile: An 800GB Dataset of Diverse Text for Language Modeling}",
      journal = {arXiv e-prints},
         year = 2020,
        month = dec,
          eid = {arXiv:2101.00027},
        pages = {arXiv:2101.00027},
          doi = {10.48550/arXiv.2101.00027},
archivePrefix = {arXiv},
       eprint = {2101.00027},
 primaryClass = {cs.CL},
       adsurl = {https://ui.adsabs.harvard.edu/abs/2021arXiv210100027G}
}

@inproceedings{henderson2022pile,
  title={Pile of Law: Learning Responsible Data Filtering from the Law and a 256GB Open-Source Legal Dataset},
  author={Henderson, Peter and Krass, Mark S. and Zheng, Lucia and Guha, Neel and Manning, Christopher D. and Jurafsky, Dan and Ho, Daniel E.},
  booktitle={Advances in Neural Information Processing Systems},
  year={2022}
}

@article{li2017learning,
  title={Learning without forgetting},
  author={Li, Zhizhong and Hoiem, Derek},
  journal={IEEE transactions on pattern analysis and machine intelligence},
  volume={40},
  number={12},
  pages={2935--2947},
  year={2017},
  publisher={IEEE}
}

@inproceedings{finn2017model,
  title={Model-agnostic meta-learning for fast adaptation of deep networks},
  author={Finn, Chelsea and Abbeel, Pieter and Levine, Sergey},
  booktitle={International conference on machine learning},
  pages={1126--1135},
  year={2017},
  organization={PMLR}
}

@inproceedings{mitchell2022memory,
  title={Memory-based model editing at scale},
  author={Mitchell, Eric and Lin, Charles and Bosselut, Antoine and Manning, Christopher D and Finn, Chelsea},
  booktitle={International Conference on Machine Learning},
  pages={15817--15831},
  year={2022},
  organization={PMLR}
}

@inproceedings{hardttest,
  title={Test-Time Training on Nearest Neighbors for Large Language Models},
  year={2024},
  author={Hardt, Moritz and Sun, Yu},
  booktitle={The Twelfth International Conference on Learning Representations},
}

@article{mitchell2021fast,
  title={Fast model editing at scale},
  author={Mitchell, Eric and Lin, Charles and Bosselut, Antoine and Finn, Chelsea and Manning, Christopher D},
  journal={arXiv preprint arXiv:2110.11309},
  year={2021}
}

@article{lopez2017gradient,
  title={Gradient episodic memory for continual learning},
  author={Lopez-Paz, David and Ranzato, Marc'Aurelio},
  journal={Advances in neural information processing systems},
  volume={30},
  year={2017}
}

@inproceedings{gururangan2020dontstop,
  title={Don’t Stop Pretraining: Adapt Language Models to Domains and Tasks},
  author={Gururangan, Suchin and Marasovi{\'c}, Ana and Swayamdipta, Swabha and Lo, Kyle and Beltagy, Iz and Downey, Doug and Smith, Noah A},
  booktitle={Proceedings of the 58th Annual Meeting of the Association for Computational Linguistics},
  pages={8342--8360},
  year={2020}
}

@article{meng2022mass,
  title={Mass-editing memory in a transformer},
  author={Meng, Kevin and Sharma, Arnab Sen and Andonian, Alex and Belinkov, Yonatan and Bau, David},
  journal={arXiv preprint arXiv:2210.07229},
  year={2022}
}

@article{meng2022locating,
  title={Locating and editing factual associations in gpt},
  author={Meng, Kevin and Bau, David and Andonian, Alex and Belinkov, Yonatan},
  journal={Advances in neural information processing systems},
  volume={35},
  pages={17359--17372},
  year={2022}
}

@article{kirkpatrick2017overcoming,
  title={Overcoming catastrophic forgetting in neural networks},
  author={Kirkpatrick, James and Pascanu, Razvan and Rabinowitz, Neil and Veness, Joel and Desjardins, Guillaume and Rusu, Andrei A and Milan, Kieran and Quan, John and Ramalho, Tiago and Grabska-Barwinska, Agnieszka and others},
  journal={Proceedings of the national academy of sciences},
  volume={114},
  number={13},
  pages={3521--3526},
  year={2017},
  publisher={National Academy of Sciences}
}

@inproceedings{zheng2021does,
  title={When does pretraining help? assessing self-supervised learning for law and the casehold dataset of 53,000+ legal holdings},
  author={Zheng, Lucia and Guha, Neel and Anderson, Brandon R and Henderson, Peter and Ho, Daniel E},
  booktitle={Proceedings of the eighteenth international conference on artificial intelligence and law},
  pages={159--168},
  year={2021}
}

@article{tsatsaronis2015overview,
  title={An overview of the BIOASQ large-scale biomedical semantic indexing and question answering competition},
  author={Tsatsaronis, George and others},
  journal={BMC Bioinformatics},
  volume={16},
  number={1},
  pages={138},
  year={2015},
  publisher={BioMed Central}
}

@article{guha2023legalbench,
  title={Legalbench: A collaboratively built benchmark for measuring legal reasoning in large language models},
  author={Guha, Neel and Nyarko, Julian and Ho, Daniel and R{\'e}, Christopher and Chilton, Adam and Chohlas-Wood, Alex and Peters, Austin and Waldon, Brandon and Rockmore, Daniel and Zambrano, Diego and others},
  journal={Advances in neural information processing systems},
  volume={36},
  pages={44123--44279},
  year={2023}
}

@ARTICLE{2025arXiv250503970Z,
       author = {{Zheng}, Lucia and {Guha}, Neel and {Arifov}, Javokhir and {Zhang}, Sarah and {Skreta}, Michal and {Manning}, Christopher D. and {Henderson}, Peter and {Ho}, Daniel E.},
        title = "{A Reasoning-Focused Legal Retrieval Benchmark}",
      journal = {arXiv e-prints},
         year = 2025,
        month = may,
          eid = {arXiv:2505.03970},
        pages = {arXiv:2505.03970},
          doi = {10.48550/arXiv.2505.03970},
archivePrefix = {arXiv},
       eprint = {2505.03970},
 primaryClass = {cs.CL},
       adsurl = {https://ui.adsabs.harvard.edu/abs/2025arXiv250503970Z}
}

@article{lazaridou2021mind,
  title={Mind the gap: Assessing temporal generalization in neural language models},
  author={Lazaridou, Angeliki and Kuncoro, Adhi and Gribovskaya, Elena and Agrawal, Devang and Liska, Adam and Terzi, Tayfun and Gimenez, Mai and de Masson d'Autume, Cyprien and Kocisky, Tomas and Ruder, Sebastian and others},
  journal={Advances in Neural Information Processing Systems},
  volume={34},
  pages={29348--29363},
  year={2021}
}

@inproceedings{petroni2019language,
  title={Language models as knowledge bases?},
  author={Petroni, Fabio and Rockt{\"a}schel, Tim and Riedel, Sebastian and Lewis, Patrick and Bakhtin, Anton and Wu, Yuxiang and Miller, Alexander},
  booktitle={Proceedings of the 2019 conference on empirical methods in natural language processing and the 9th international joint conference on natural language processing (EMNLP-IJCNLP)},
  pages={2463--2473},
  year={2019}
}

@article{sheng2023s,
  title={S-lora: Serving thousands of concurrent lora adapters},
  author={Sheng, Ying and Cao, Shiyi and Li, Dacheng and Hooper, Coleman and Lee, Nicholas and Yang, Shuo and Chou, Christopher and Zhu, Banghua and Zheng, Lianmin and Keutzer, Kurt and others},
  journal={arXiv preprint arXiv:2311.03285},
  year={2023}
}

@article{chen2024punica,
  title={Punica: Multi-tenant lora serving},
  author={Chen, Lequn and Ye, Zihao and Wu, Yongji and Zhuo, Danyang and Ceze, Luis and Krishnamurthy, Arvind},
  journal={Proceedings of Machine Learning and Systems},
  volume={6},
  pages={1--13},
  year={2024}
}

@inproceedings{gururangan2020don,
  title={Don’t stop pretraining: Adapt language models to domains and tasks},
  author={Gururangan, Suchin and Marasovi{\'c}, Ana and Swayamdipta, Swabha and Lo, Kyle and Beltagy, Iz and Downey, Doug and Smith, Noah A},
  booktitle={Proceedings of the 58th annual meeting of the association for computational linguistics},
  pages={8342--8360},
  year={2020}
}

@article{jang2021towards,
  title={Towards continual knowledge learning of language models},
  author={Jang, Joel and Ye, Seonghyeon and Yang, Sohee and Shin, Joongbo and Han, Janghoon and Kim, Gyeonghun and Choi, Stanley Jungkyu and Seo, Minjoon},
  journal={arXiv preprint arXiv:2110.03215},
  year={2021}
}

@inproceedings{tu2025robust,
  title={Robust Fine-tuning for Retrieval Augmented Generation against Retrieval Defects},
  author={Tu, Yiteng and Su, Weihang and Zhou, Yujia and Liu, Yiqun and Ai, Qingyao},
  booktitle={Proceedings of the 48th International ACM SIGIR Conference on Research and Development in Information Retrieval},
  pages={1272--1282},
  year={2025}
}

@inproceedings{wang2026joint,
  title={Joint evaluation of answer and reasoning consistency for hallucination detection in large reasoning models},
  author={Wang, Changyue and Su, Weihang and Ai, Qingyao and Liu, Yiqun},
  booktitle={Proceedings of the AAAI Conference on Artificial Intelligence},
  volume={40},
  number={39},
  pages={33377--33385},
  year={2026}
}

@article{su2025towards,
  title={Towards Unification of Hallucination Detection and Fact Verification for Large Language Models},
  author={Su, Weihang and Long, Jianming and Wang, Changyue and Lin, Shiyu and Xu, Jingyan and Ye, Ziyi and Ai, Qingyao and Liu, Yiqun},
  journal={arXiv preprint arXiv:2512.02772},
  year={2025}
}

@inproceedings{wang2025decoupling-2,
  title={Decoupling reasoning and knowledge injection for in-context knowledge editing},
  author={Wang, Changyue and Su, Weihang and Ai, Qingyao and Zhou, Yujia and Liu, Yiqun},
  booktitle={Findings of the Association for Computational Linguistics: ACL 2025},
  pages={24543--24562},
  year={2025}
}

@inproceedings{wang2025knowledge,
  title={Knowledge editing through chain-of-thought},
  author={Wang, Changyue and Su, Weihang and Ai, Qingyao and Tang, Yichen and Liu, Yiqun},
  booktitle={Proceedings of the 2025 Conference on Empirical Methods in Natural Language Processing},
  pages={10684--10704},
  year={2025}
}

@article{su2025pre,
  title={Pre-training for legal case retrieval based on inter-case distinctions},
  author={Su, Weihang and Ai, Qingyao and Wu, Yueyue and Xie, Anzhe and Wang, Changyue and Ma, Yixiao and Li, Haitao and Wu, Zhijing and Liu, Yiqun and Zhang, Min},
  journal={ACM Transactions on Information Systems},
  volume={43},
  number={5},
  pages={1--27},
  year={2025},
  publisher={ACM New York, NY}
}

@article{su2026enhancing,
  title={Enhancing Judgment Document Generation via Agentic Legal Information Collection and Rubric-Guided Optimization},
  author={Su, Weihang and Chen, Xuanyi and Wu, Yueyue and Ai, Qingyao and Liu, Yiqun},
  journal={arXiv preprint arXiv:2605.02011},
  year={2026}
}

@inproceedings{su2024wikiformer,
  title={Wikiformer: Pre-training with structured information of wikipedia for ad-hoc retrieval},
  author={Su, Weihang and Ai, Qingyao and Li, Xiangsheng and Chen, Jia and Liu, Yiqun and Wu, Xiaolong and Hou, Shengluan},
  booktitle={Proceedings of the AAAI Conference on Artificial Intelligence},
  volume={38},
  number={17},
  pages={19026--19034},
  year={2024}
}

@inproceedings{su2025sigirap,
  title={SIGIR-AP 2025 Tutorial Proposal: Dynamic and Parametric Retrieval-Augmented Generation},
  author={Su, Weihang and Dong, Qian and Ai, Qingyao and Liu, Yiqun},
  booktitle={3rd International ACM SIGIR Conference on Information Retrieval in the Asia Pacific},
  year={2025}
}

@article{wang2026adaptive,
  title={Adaptive Multi-Resolution Procedural Knowledge Compression for Large Language Models},
  author={Wang, Changyue and Su, Weihang and Ai, Qingyao and Tang, Yichen and Qiao, Runzhong and Li, Xuancheng and Zhang, Min and Liu, Yiqun},
  journal={arXiv preprint arXiv:2606.12203},
  year={2026}
}

@article{su2025surge,
  title={Surge: A benchmark and evaluation framework for scientific survey generation},
  author={Su, Weihang and Xie, Anzhe and Ai, Qingyao and Long, Jianming and Chen, Xuanyi and Mao, Jiaxin and Ye, Ziyi and Liu, Yiqun},
  journal={arXiv preprint arXiv:2508.15658},
  year={2025}
}

@article{su2026decoupling,
  title={Decoupling Knowledge and Task Subspaces for Composable Parametric Retrieval Augmented Generation},
  author={Su, Weihang and Zhang, Hanwen and Ai, Qingyao and Liu, Yiqun},
  journal={arXiv preprint arXiv:2604.26768},
  year={2026}
}

@article{jin2025search,
  title={Search-r1: Training llms to reason and leverage search engines with reinforcement learning},
  author={Jin, Bowen and Zeng, Hansi and Yue, Zhenrui and Yoon, Jinsung and Arik, Sercan and Wang, Dong and Zamani, Hamed and Han, Jiawei},
  journal={arXiv preprint arXiv:2503.09516},
  year={2025}
}

@article{su2026skill,
  title={Skill Retrieval Augmentation for Agentic AI},
  author={Su, Weihang and Long, Jianming and Ai, Qingyao and Tang, Yichen and Wang, Changyue and Tu, Yiteng and Liu, Yiqun},
  journal={arXiv preprint arXiv:2604.24594},
  year={2026}
}

\appendix

\section{Additional Discussion of Related Work}\label{app:related}

\subsection{Continual Post-Training and Knowledge Editing.}
Continued post-training and fine-tuning are standard paradigms for injecting new knowledge into LLMs \citep{gururangan2020dontstop}.
A core challenge in these paradigms is catastrophic forgetting and capability drift under long sequences of updates, motivating the use of mitigation techniques in continual learning, such as regularization, distillation, or replay \citep{kirkpatrick2017overcoming,li2017learning,lopez2017gradient}.
In parallel, knowledge editing aims to precisely update facts with minimal side effects \citep{meng2022locating}.
Representative editors include localized weight updates \citep{meng2022mass}, meta-learned editors that transform gradients into parameter shifts \citep{mitchell2021fast}, and memory-based editors \citep{mitchell2022memory}.

\subsection{Retrieval-Augmented Generation}
\label{sec:rw_rag}

Retrieval-Augmented Generation (RAG) has become a widely used paradigm for equipping LLMs with external knowledge without modifying their parameters~\cite{lewis2020retrieval,dong2025decoupling,tu2025robust,su2025dynamic,su2025surge}.
Instead of internalizing new information through additional training, RAG retrieves query-relevant passages at inference time and provides them to the model as part of the input context.
This non-parametric design makes RAG particularly attractive for hallucination mitigation~\cite{wang2026joint,su2025towards,su2024unsupervised,wang2025decoupling-2}, knowledge updates~\cite{wang2025decoupling,wang2025knowledge,wang2024lekube}, and domains adaption without full model retraining~\cite{su2024stard,su2025judge,su2025pre,su2026enhancing,su2023caseformer}.
A large body of work has improved different components of the RAG pipeline, including sparse and dense retrieval~\cite{robertson2009probabilistic,su2024wikiformer,fang2024scaling}, adaptive retrieval strategies~\cite{jiang2023active,su2024dragin,su2024mitigating}, graph-structured retrieval~\cite{edge2024local}, parametric RAG~\cite{su2025parametric,su2025sigirap,su2026decoupling,wang2026adaptive}, and agentic retrieval frameworks~\cite{jin2025search,su2026skill}.
\textsc{ReGrad} shares the retrieval-based philosophy of RAG, but changes what is retrieved and how the retrieved information is used.

\subsection{Test-Time Training.}
Recent works explore training the model at inference time, e.g., test-time fine-tuning on retrieved neighbors \citep{hardttest,chen2023reckoning}.
Such methods demonstrate that a few gradient steps can yield strong specialization, but require online backpropagation and training, which will introduce significant latency.
ReGrad bypasses the latency constraints of online backpropagation by retrieving pre-computed, meta-optimized parameter shifts. This approach decouples optimization from inference, thereby retaining the benefits of instance-specific adaptation while incurring negligible runtime overhead.

\section{Additional Robustness Analysis}
\label{app:noise}

\begin{figure}[t]
\centering
\includegraphics[width=\columnwidth]{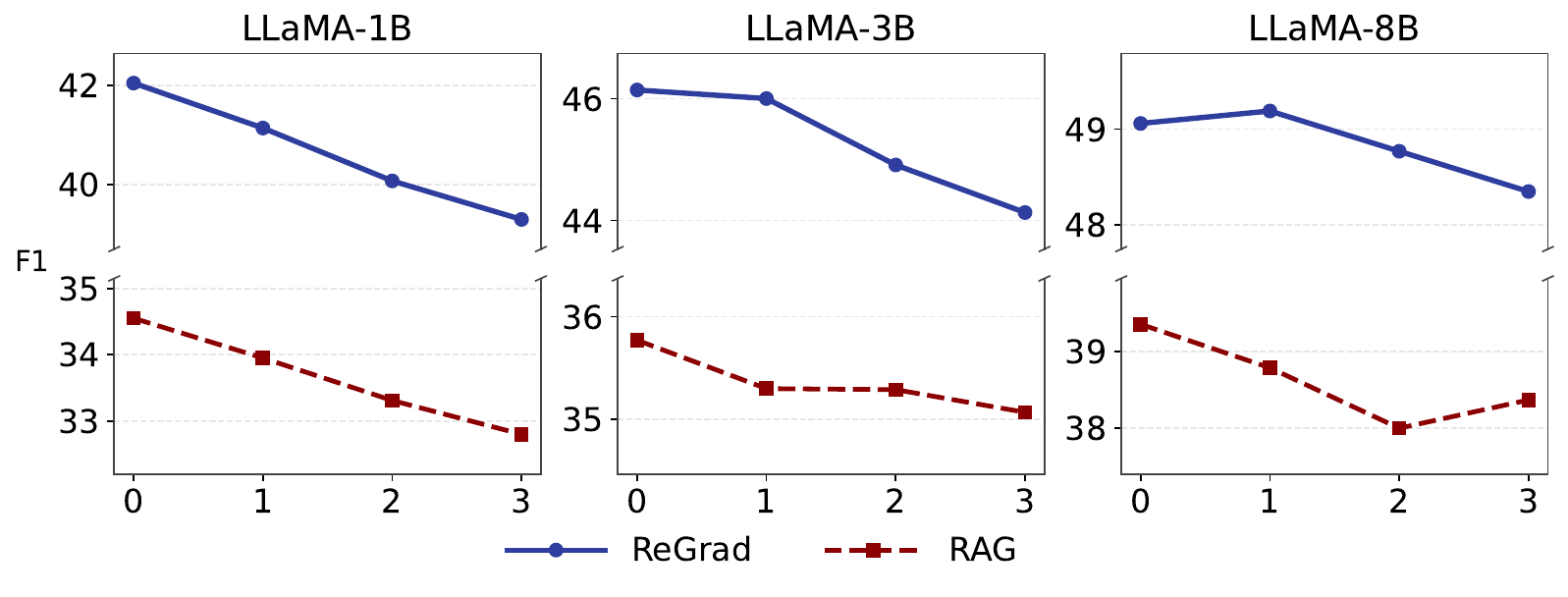}
\caption{Average performance across four datasets with varying levels of noise ($3$ relevant documents plus $n$ random documents). \textsc{ReGrad} maintains a consistent performance advantage over RAG across all model scales despite similar degradation trends.}
\label{fig:noise}
\end{figure}

In Section~\ref{sec:merge}, we analyze how \textsc{ReGrad} behaves when merging gradients from an increasing number of retrieved documents.
Here, we provide a complementary stress test that explicitly isolates the effect of noisy retrieval.
Specifically, we keep a fixed set of $k=3$ relevant passages and inject $n \in \{0, 1, 2, 3\}$ random distractor documents.
As illustrated in Figure~\ref{fig:noise}, \textsc{ReGrad} consistently outperforms RAG across all noise levels and model sizes, demonstrating that our parametric adaptation yields stronger knowledge utilization even in the presence of retrieval noise.
However, we also observe that \textsc{ReGrad} exhibits a performance degradation pattern parallel to that of RAG as noise increases.
This indicates that while \textsc{ReGrad} effectively capitalizes on relevant information, it remains sensitive to irrelevant signals introduced by noisy retrieval.
This finding points to important directions for future optimization: enhancing robustness not only through upstream retrieval denoising, but also by exploring more selective mechanisms for merging knowledge atoms beyond the current accumulation strategy.

\section{Detailed Discussion on Time and Space Efficiency}
\label{sec:app-efficiency}

We analyze the computational and storage cost of \textsc{ReGrad}, with a particular focus on how it compares with continual post-training and RAG. 

\subsection{Computation Cost}
\label{sec:efficiency-compute}

\paragraph{Meta-training overhead.}
The meta-learning stage optimizes the initialization $\theta$ and step sizes $\alpha$ such that unsupervised document gradients become useful for downstream QA.
In practice, this stage incurs a relatively small one-time cost because the meta-objective converges on only a few thousand training instances, and the total computation is negligible compared to corpus-scale processing.
We therefore focus our discussion on the dominant cost: gradient construction over the knowledge corpus.

\paragraph{Offline gradient construction.}
For each document $d_i \in \mathcal{D}$, \textsc{ReGrad} computes a single backward pass of the language modeling loss restricted to the low-rank plug-in parameters $\theta$.
Crucially, this gradient is computed \emph{once} and stored in the Gradient Bank.
As a result, the total offline cost is equivalent to performing one epoch of continual post-training over the corpus, except that updates are confined to a small LoRA subspace rather than the full model.
Compared to CPT, which typically requires multiple epochs to integrate corpus-level knowledge, \textsc{ReGrad} reduces the total training computation by a factor proportional to the number of CPT epochs.
If CPT performs $n$ epochs over $\mathcal{D}$, then \textsc{ReGrad} incurs approximately $\frac{1}{n}$ of the total optimization cost, while avoiding cumulative parameter drift.

\paragraph{Online inference cost.}
At inference time, \textsc{ReGrad} retrieves a small set of gradients $\{g_{i}\}_{i=1}^{t}$ and aggregates them into a temporary update applied to the plug-in parameters.
This operation consists of simple tensor additions and scalar scaling in the low-rank space and is computationally negligible compared to even a single token-level forward pass.

Let $|q|$ denote the query length and $h$ the hidden dimension.
The inference-time complexity of \textsc{ReGrad} remains
\begin{equation}
    \mathcal{O}(|q|^2 h + |q| h^2),
\end{equation}

\noindent identical to that of the base LLM.
In contrast, in-context RAG expands the input by concatenating retrieved documents, yielding a complexity of
\begin{equation}
\mathcal{O}\big((|q| + t|d|)^2 h + (|q| + t|d|) h^2\big),
\end{equation}

\noindent which grows quadratically with the total retrieved context length.
Therefore, \textsc{ReGrad} strictly outperforms in-context RAG in online efficiency: it achieves query-conditioned knowledge activation without increasing context length.
The trade-off is explicit: \textsc{ReGrad} shifts computation from the online phase to an offline, amortizable preprocessing stage.

\paragraph{When is \textsc{ReGrad} more compute-efficient?} 
Consider a corpus of $100$k documents, each containing $|d|$ tokens, and an RAG system that retrieves $t=3$ documents per query. We quantify computational cost in units of token-wise forward passes. In this analysis, we omit the costs of RAG indexing and \textsc{ReGrad}'s Meta-Learning Initialization phase, as they are negligible compared to the construction of the Gradient Bank. Additionally, since \textsc{ReGrad} stores LoRA gradients, the computation required to merge or insert gradients is significantly lower than a single token forward pass through the full LLM and is therefore disregarded. 

Based on these assumptions, we now perform a quantitative comparison of the computational overheads. Constructing the Gradient Bank requires $100$k backward passes; assuming the computational cost of a backward pass is approximately three times that of a forward pass, the total offline pre-computation is equivalent to forward-propagating $300\text{k} \cdot |d|$ tokens. In contrast, RAG incurs an additional forward computation overhead proportional to $3|d|$ tokens for \emph{every} query. Consequently, once the volume of served queries exceeds the corpus size (i.e., $100$k queries), the amortized cost of \textsc{ReGrad} becomes lower than that of RAG. This regime is characteristic of long-lived knowledge bases or high-traffic QA systems, where offline preprocessing costs are effectively amortized across a large number of queries.

\subsection{Storage Overhead}
\label{sec:efficiency-storage}

The storage cost of \textsc{ReGrad} arises from storing per-document gradients in the plug-in parameter space.
Let $r$ denote the LoRA rank, $n$ the number of Transformer layers, $h$ the hidden dimension, and $l$ the FFN intermediate dimension.
Each document gradient contains $2nr(h+l)$ parameters.
While this is larger than storing raw text alone, the representation is compact relative to full-model checkpoints and scales linearly with the number of documents.
In practice, this storage overhead is manageable for several reasons.
First, access patterns in real-world systems are highly skewed: a small fraction of documents accounts for the majority of queries.
Maintaining gradient representations only for high-frequency or high-value documents already captures most of the benefit.
Furthermore, when storage is constrained, gradients need not be permanently stored.
They can be computed on the fly at inference time using the fixed $(\theta^*, \alpha^*)$.
As shown in our experiments, online gradient computation introduces only millisecond-level latency, making it feasible for interactive settings.
This flexibility allows practitioners to trade storage for computation depending on system constraints.

\begin{table}[t]
\centering
\caption{
End-to-end inference latency in milliseconds.
\textsc{ReGrad} Cached loads pre-computed gradients from the Gradient Bank, while \textsc{ReGrad} Online computes document gradients on the fly without storing cached gradients.
}
\label{tab:app_latency}
\setlength{\tabcolsep}{6pt}
\resizebox{0.94\linewidth}{!}{
\begin{tabular}{lccc}
\toprule
\textbf{Method} & \textbf{1B} & \textbf{3B} & \textbf{8B} \\
\midrule
Direct Generation & 13.44 ms & 18.36 ms & 28.30 ms \\
RAG & 17.03 ms& 23.62 ms& 34.63 ms\\
\textsc{ReGrad} Cached & 16.89 ms& 22.97 ms& 30.84 ms\\
\textsc{ReGrad} + ICL & 18.21 ms& 26.60 ms& 38.10 ms\\
\textsc{ReGrad} Online & 32.04 ms& 45.83 ms& 56.31 ms\\
\bottomrule
\end{tabular}
}
\end{table}

\section{End-to-End Latency }
\label{sec:app-latency}

We further conduct an end-to-end latency evaluation to quantify the practical cost of dynamic low-rank update merging and to examine the storage-latency trade-off of \textsc{ReGrad}.
All experiments are conducted on an 8$\times$A100-80G server.
We report the average latency over 10 repeated runs on 100 randomly sampled general-domain questions.
Table~\ref{tab:app_latency} shows that the additional serving overhead of cached \textsc{ReGrad} is small.
For the 8B model, Direct Generation takes 28.30 ms, while \textsc{ReGrad} Cached takes 30.84 ms.
This confirms that dynamic low-rank update merging introduces negligible latency overhead in practice.
We also evaluate \textsc{ReGrad} Online, an extreme setting that does not store a cached Gradient Bank.
Instead, the document gradients are computed on the fly at inference time.
This removes the storage cost of cached gradients, while retaining the raw document corpus and retrieval index.
As expected, this setting is slower than cached \textsc{ReGrad}, but the overhead remains moderate.
For the 8B model, \textsc{ReGrad} Online takes 56.31 ms, which is 25.47 ms higher than cached \textsc{ReGrad}.
This result suggests a practical hybrid deployment strategy:
systems can cache gradients for frequently retrieved head documents to maximize inference speed, while computing gradients for long-tail documents on demand to reduce storage usage.

Overall, these results complement the theoretical efficiency analysis by showing that \textsc{ReGrad} supports a flexible latency--storage trade-off.
Cached gradients provide near-RAG-level latency with negligible dynamic merging cost, whereas on-the-fly gradient computation offers a storage-saving alternative when full Gradient Bank caching is undesirable.

\section{Experimental Setup}

In this section, we detail the experimental settings used to evaluate our proposed  ReGrad framework. 
We begin by introducing our selected benchmark datasets and evaluation metrics (\S\ref{app:benchmarks} and \S\ref{app:metrics}). 
Next, we introduce our selected baseline methods (\S\ref{app:baselines}) and implementation. 
Finally, we provide implementation details regarding our meta-learning process, retrieval strategy, and inference settings (\S\ref{app:implementation}).

\subsection{Baselines}\label{app:baselines}

We compare against a broad set of baselines, including direct generation (no external knowledge provided), parametric knowledge injection (Standard CPT and its variant Instruction CPT~\cite{cheng2023adapting}), non-parametric knowledge injection (Standard RAG~\citep{lewis2020retrieval}), and a retrieval-based parameter-update method (Parametric RAG~\citep{su2025parametric}). Below, we introduce each baseline and present its implementation details.

\begin{itemize}[leftmargin=*]

\item \textbf{Direct Generation:} This baseline evaluates the backbone LLM's intrinsic capabilities on downstream tasks without any external knowledge injection. We use the official Hugging Face inference implementation and follow the standard generation configurations provided by the model developers. The specific prompt templates used for zero-shot inference are detailed in Appendix~\ref{app:prompt}.

\item \textbf{Standard Continual Post-Training (Standard CPT):} Representing the conventional domain adaptation approach, we update the model parameters on the external corpus using a standard causal language modeling objective. 
To avoid diluting this baseline with a large number of irrelevant documents, we construct its training corpus from the top-3 documents retrieved for each question in the training set, rather than using the entire external corpus.
We then perform LoRA fine-tuning on the external knowledge corpus for $1$ epoch using the standard language modeling objective. We utilize a learning rate of $5e^{-5}$ and a batch size of $16$. The inference configuration remains consistent with Direct Generation.

\item \textbf{Instruction Continual Post-Training (Instruction CPT)}~\citep{cheng2023adapting}: To mitigate the potential loss of reasoning capabilities during domain adaptation, we adopt the core philosophy of this framework, which transforms raw corpora into reading comprehension tasks. However, our implementation diverges from~\citet{cheng2023adapting} in the data construction pipeline. While the original method relies on unsupervised rule matching to extract Question-Answer (QA) pairs, our preliminary experiments indicate that this approach yields suboptimal performance. Consequently, we replace the rule-based heuristic with an LLM-based generation approach, prompting the model to synthesize high-quality QA pairs for each document (templates provided in Appendix~\ref{app:prompt}). We then post-train the model using a language modeling objective on the concatenated sequence of the document context and the generated QA pair. The training hyperparameters align with Standard CPT (learning rate $1e^{-5}$, batch size $16$, $1$ epoch).

\item \textbf{Standard RAG:} We employ the conventional Retrieval-Augmented Generation framework, which serves as a strong non-parametric baseline. Following the settings in the PRAG library, we retrieve the top-$3$ relevant documents from the external corpus and prepend them to the input prompt as context. This allows the model to use external information through in-context learning without modifying its weights. The prompt templates are aligned with those used in ReGrad and are listed in Appendix~\ref{app:prompt}.

\item \textbf{Parametric RAG}~\citep{su2025parametric}: This recent approach bridges the gap between RAG and CPT by injecting relevant documents directly into the LLM’s parameters via document parameterization, rather than consuming context window space. We utilize the official implementation from the PRAG library to compare ReGrad against state-of-the-art retrieval-based parameter-update methods.

\item \textbf{Finetuned Direct Generation:} To ensure a fair comparison with ReGrad's meta-learning phase, we implement a supervised fine-tuning (SFT) baseline. We fine-tune the base LLM on QA training sets that have no overlap with the corresponding test sets, producing separate models for the General, Law, and Medical domains. 
The model is trained to generate the answer given the question, with the language modeling loss calculated solely on the answer tokens. We employ LoRA for efficient fine-tuning, setting the learning rate to $4e^{-4}$ for $1$ epoch.

\item \textbf{Finetuned Standard RAG:} We fine-tune the base LLM using the same hyperparameters as Finetuned Direct Generation (LoRA, learning rate $4e^{-4}$ for $1$ epoch). Crucially, unlike the Direct Generation baseline, we retrieve the top-3 relevant documents and append them to the input prompt during both training and inference. This aligns the training distribution with the retrieval-augmented inference setting.

\item \textbf{Finetuned Parametric RAG:} Similar to Finetuned Standard RAG, this baseline fine-tunes the model on downstream QA pairs. However, instead of appending documents to the context window, it utilizes the Parametric RAG mechanism to inject the top-$3$ retrieved documents into the model parameters during the forward pass. All training hyperparameters remain consistent with those of the other fine-tuned baselines.

\end{itemize}

To facilitate reproducibility, we have made our full implementation publicly available. \textbf{Further details regarding the setting, hyperparameters, code, data, and models can be accessed via our GitHub repository}\footnote{We have open-sourced all the code, data, and models in the following anonymized GitHub link: \url{https://github.com/oneal2000/ReGrad/}}.

\subsection{Implementation Details}
\label{app:implementation}

\textbf{Meta-Learning Configuration.}
We optimize the bi-level objective (Eq.~\ref{eq:meta_objective}) for $1$ epoch with a meta-batch size of $16$. The outer optimization employs a linear learning rate scheduler with a peak learning rate of $4e^{-4}$. {Specifically, we implement the gradient aggregation in Eq.~\ref{eq:aggregation} using top-$K$ retrieved passages with $K=3$.} To accommodate these inputs, we set the maximum context length to $1,536$ tokens. All meta-training experiments use bfloat16 mixed precision. Regarding initialization, the LoRA parameters $\theta$ follow the standard initialization from \citet{hulora}, while the learnable scalar step sizes $\alpha$ are initialized to $1e^{-2}$, providing a stable starting point for the inner loop updates.

\textbf{Retrieval and Indexing.}
While dense retrieval is popular, recent analyses in RAG~\citep{ram2023context} suggest that lexical matching remains highly competitive, particularly for zero-shot generalization. Given its efficiency and proven robustness, we adopt {BM25} as our retrieval backbone, implemented via Elasticsearch. This ensures that performance gains in ReGrad are attributed to the gradient injection mechanism rather than the retrieval of superior contexts.

\textbf{Base Models and Computing Environment.}
We implement ReGrad using the open-source LLaMA-3.2 family to verify scalability and robustness across varying model sizes. Specifically, we conduct experiments on {LLaMA-3.2-1B-Instruct}, {LLaMA-3.2-3B-Instruct}, and {LLaMA-3.1-8B-Instruct}. All training and inference processes are executed using PyTorch on NVIDIA A100 GPUs (80GB).

\textbf{Inference Configuration.}
All evaluations utilize the official Hugging Face implementations of the LLaMA models. We adhere to the default hyperparameters and chat templates provided by the library. To strictly ensure the reproducibility of our reported metrics, we employ greedy decoding (temperature $= 0$) for all generation tasks.

\subsection{Benchmarks and Corpus}
\label{app:benchmarks}

\paragraph{General Domain.} Consistent with standard practice in retrieval-augmented generation~\citep{karpukhin2020dense,su2024dragin}, we utilize the  Wikipedia dump provided by DPR~\citep{karpukhin2020dense} as our external knowledge corpus. To evaluate reasoning capabilities and knowledge retrieval over this corpus, we select four diverse benchmarks: 

\begin{itemize}[leftmargin=*] 

\item \textbf{2WikiMultihopQA (2WQA)}~\citep{ho2020constructing}: Evaluates multi-hop reasoning by requiring the integration of information across fragmented Wikipedia passages. 

\item \textbf{HotpotQA (HQA)}~\citep{yang2018hotpotqa}: A challenging benchmark for multi-hop QA that necessitates reasoning over multiple supporting documents to derive the answer. 

\item \textbf{PopQA (PQA)}~\citep{mallen-etal-2023-trust}: Assesses factual knowledge retrieval, specifically targeting the resolution of entity ambiguity and long-tail knowledge. 

\item \textbf{ComplexWebQuestions (CWQ)}~\citep{talmor-berant-2018-web}: Requires answering complex, multi-step questions derived from the web, testing the scalability of the retrieval mechanism. 
\end{itemize}

\paragraph{Vertical Domains.} To verify the effectiveness of ReGrad in specialized fields, we conduct experiments in Medicine and Law. For each domain, we utilize a specific uncopyrighted corpus to construct the Gradient Bank.

\textbf{Medicine:} We utilize the \textbf{PubMed Abstracts} subset of The Pile~\citep{2021arXiv210100027G} as the reference corpus. Evaluation is performed on three datasets: 

\begin{itemize}[leftmargin=*, nosep] 

\item \textbf{PubMedQA}~\citep{jin2019pubmedqa}: A biomedical QA task derived from PubMed abstracts, requiring the model to deduce research conclusions (Yes/No/Maybe). 

\item \textbf{MedQA}~\citep{jin2021disease}: Comprises questions from professional medical board exams, assessing complex clinical reasoning. 

\item \textbf{BioASQ}~\citep{tsatsaronis2015overview}: We utilize the binary (Yes/No) subset of the BioASQ-12b training data, where questions involve diverse biomedical topics including diseases and pharmaceuticals. 

\end{itemize}

\textbf{Law:} 
We employ the \textbf{Pile-of-Law}~\citep{henderson2022pile} as the legal reference corpus. Downstream evaluation includes: \begin{itemize}[leftmargin=*, nosep] \item \textbf{CaseHOLD}~\citep{zheng2021does}: A multiple-choice task (5 options) requiring the identification of the correct holding statement for a given legal citation context. \item \textbf{Learned Hands Family (LHF)}~\citep{guha2023legalbench}: A binary classification task determining whether a legal query pertains to family law issues (e.g., custody, divorce). \item \textbf{Housing QA}~\citep{2025arXiv250503970Z}: A binary QA dataset assessing reasoning capabilities regarding US state-level housing laws. \end{itemize}

\subsection{Evaluation Protocols and Metrics} 
\label{app:metrics}

To maintain consistency across diverse benchmarks, we evaluate models using the first 1000 instances from each sub-dataset test set. Performance is assessed via task-specific metrics. Accuracy is applied to all closed-ended classification tasks, including multiple-choice questions in CaseHOLD and binary classification in PubMedQA, BioASQ, LHF, and Housing QA; model outputs are mapped to labels using regular expression parsing. For open-ended question-answering tasks, including 2WQA, HQA, PopQA, CWQ, and MedQA, we use F1 Score to measure overlap between generated responses and ground truth. This metric ensures a robust evaluation of generation quality by penalizing partial hallucinations while rewarding precise recall of relevant information.

\section{Prompt Templates}
\label{app:prompt}

This section describes the prompt templates used across all the methods and experiment stages (training vs. validation). We offer a comprehensive account of how the context is structured, how it is incorporated into the prompt, and how this design differs between baseline approaches and our proposed methods, as well as across datasets with different question formats.

\subsection{Context Construction}

For multi-passage datasets such as {2WikiMultihopQA}, {HotpotQA}, {PopQA}, and {ComplexWebQuestions}, each example is associated with a set of retrieved passages, which provide the background knowledge to answer a multihop question. These passages are concatenated into a single context string indexed in a fixed sequential order.
Formally, the context is constructed as:
\begin{tcolorbox}[colback=lightgray!20,colframe=darkgray!80,title=Concatenated Context]
\small
Passage 0: \{Passage 0\}\\
Passage 1: \{Passage 1\}\\
...\\
Passage k-1: \{Passage k-1\}
\end{tcolorbox}
The concatenated context is truncated to a fixed maximum length to satisfy the input constraints of the language models.
In our implementation, the context is truncated to at most 1536 tokens during training and 2048 tokens during validation and inference. This enables the model to attend to the most relevant context while keeping the input length within allowable limits.

\subsection{Prompt Usage Across Methods}

The prompt templates mainly differ in whether the retrieved relevant document is included in the context, and are further adapted across datasets with different question formats.
We describe below how each method uses the prompt and whether the context is available during answer generation.

\textbf{RAG},  \textbf{Fine-tuned-RAG} and \textbf{ReGrad + ICL} use prompts with context. In these settings, the relevant background information is directly exposed to the model, which is expected to exploit it during answer generation.
\textbf{Direct}, \textbf{Standard CPT}, \textbf{Instruction CPT}, \textbf{PRAG}, \textbf{Fine-tuned}, \textbf{Fine-tuned-PRAG} and \textbf{ReGrad} use prompts without context during training and validation, requiring the model to rely on parametric knowledge or learned plug-in parameters.

\subsection{Prompt Templates}

We present all our prompt templates in this subsection.
Depending on the dataset type, slight variations are applied to open-ended, yes/no, and multiple-choice QA. For yes/no datasets (e.g., {MedQA},{PubMedQA}, {BioASQ}, {LHF}, {HousingQA}), the same templates are used with task instructions adjusted to enforce binary outputs. 

\vspace{5mm}
\textbf{Prompt with context for open-ended QA:}
This template is used for open-ended question answering tasks when contextual passages are provided:
\begin{tcolorbox}[colback=lightgray!20,colframe=darkgray!80,title=Prompt with context for open-ended QA, before skip=5mm, after skip=5mm]
\small
\tk{begin\_of\_text}\tk{start\_header\_id}system\tk{end\_header\_id}\\
\#\#\#\# CONTEXT begin \#\#\#\#\\
\{context\}\\
\#\#\#\# CONTEXT end \#\#\#\#\\
You are a helpful AI assistant who is ready to answer user's question according to the CONTEXT.\\
Your answer should be concise, which means it must be a short phrase or a single word, and sentences are not allowed!\\[2mm]
\tk{eot\_id}\tk{start\_header\_id}user\tk{end\_header\_id}\\
\{question\}\\[2mm]
\tk{eot\_id}\tk{start\_header\_id}assistant\tk{end\_header\_id}\\
Answer:
\end{tcolorbox}
\textbf{Prompt with context for yes/no QA:}
This template is used for binary (yes/no) question answering tasks when contextual passages are provided:
\begin{tcolorbox}[colback=lightgray!20,colframe=darkgray!80,title=Prompt with context for yes/no QA, before skip=5mm, after skip=5mm]
\small
\tk{begin\_of\_text}\tk{start\_header\_id}system\tk{end\_header\_id}\\
\#\#\#\# CONTEXT begin \#\#\#\#\\
\{context\}\\
\#\#\#\# CONTEXT end \#\#\#\#\\
You are a helpful AI assistant who is ready to answer user's question according to the CONTEXT.\\
Your answer should be concise, which means it must be 'yes' or 'no' only!\\[2mm]
\tk{eot\_id}\tk{start\_header\_id}user\tk{end\_header\_id}\\
\{question\}\\[2mm]
\tk{eot\_id}\tk{start\_header\_id}assistant\tk{end\_header\_id}\\
Answer:
\end{tcolorbox}

\textbf{Prompt with context for multiple-choice QA:}
This template is used for multiple-choice question answering tasks when contextual passages are provided:

\begin{tcolorbox}[colback=lightgray!20,colframe=darkgray!80,title=Prompt with context for multiple-choice QA, before skip=5mm, after skip=5mm]
\small
\tk{begin\_of\_text}\tk{start\_header\_id}system\tk{end\_header\_id}\\
\#\#\#\# CONTEXT begin \#\#\#\#\\
\{context\}\\
\#\#\#\# CONTEXT end \#\#\#\#\\
Complete the following excerpt from a US court opinion.\\
Always answer ONLY with the option letter: the final answer is X, where X in \{\{A, B, C, D, E\}\}.\\[2mm]
\tk{eot\_id}\tk{start\_header\_id}user\tk{end\_header\_id}\\
\{question\}\\
Choices:\\
A. \{c0\}\\
B. \{c1\}\\
C. \{c2\}\\
D. \{c3\}\\
E. \{c4\}\\[2mm]
\tk{eot\_id}\tk{start\_header\_id}assistant\tk{end\_header\_id}\\
Answer:
\end{tcolorbox}

\textbf{Prompt without context for open-ended QA:}
This template is used for open-ended question answering tasks when no contextual passages are exposed to the model:

\begin{tcolorbox}[colback=lightgray!20,colframe=darkgray!80,title=Prompt without context for open-ended QA, before skip=5mm, after skip=5mm]
\small
\tk{begin\_of\_text}\tk{start\_header\_id}system\tk{end\_header\_id}\\
You are a helpful AI assistant who is ready to answer user's question.\\
Your answer should be concise, which means it must be a short phrase or a single word, and sentences are not allowed!\\[2mm]
\tk{eot\_id}\tk{start\_header\_id}user\tk{end\_header\_id}\\
\{question\}\\[2mm]
\tk{eot\_id}\tk{start\_header\_id}assistant\tk{end\_header\_id}\\
Answer:
\end{tcolorbox}


\textbf{Prompt without context for yes/no QA:}
This template is used for binary (yes/no) question answering tasks when no contextual passages are exposed to the model:

\begin{tcolorbox}[colback=lightgray!20,colframe=darkgray!80,title=Prompt without context for yes/no QA, before skip=5mm, after skip=5mm]
\small
\tk{begin\_of\_text}\tk{start\_header\_id}system\tk{end\_header\_id}\\
You are a helpful AI assistant who is ready to answer user's question.\\
Answer only 'yes' or 'no'!\\[2mm]
\tk{eot\_id}\tk{start\_header\_id}user\tk{end\_header\_id}\\
\{question\}\\[2mm]
\tk{eot\_id}\tk{start\_header\_id}assistant\tk{end\_header\_id}\\
Answer:
\end{tcolorbox}

\textbf{Prompt without context for multiple-choice QA:}
This template is used for multiple-choice question answering tasks when no contextual passages are exposed to the model:

\begin{tcolorbox}[colback=lightgray!20,colframe=darkgray!80,title=Prompt without context for multiple-choice QA, before skip=5mm, after skip=5mm]
\small
\tk{begin\_of\_text}\tk{start\_header\_id}system\tk{end\_header\_id}\\
Complete the following excerpt from a US court opinion.\\
Answer only with the option letter: A, B, C, D or E.\\[2mm]
\tk{eot\_id}\tk{start\_header\_id}user\tk{end\_header\_id}\\
\{question\}\\
Choices:\\
A. \{c0\}\\
B. \{c1\}\\
......\\[2mm]
\tk{eot\_id}\tk{start\_header\_id}assistant\tk{end\_header\_id}\\
Answer:
\end{tcolorbox}


\textbf{Prompt for synthetic QA pairs:} This template is used for the Instruction CPT baseline, designed to instruct the LLM to generate synthetic QA pairs for each document in the corpus.

\begin{tcolorbox}[colback=lightgray!20,colframe=darkgray!80,title=Prompt for synthetic QA pairs, before skip=5mm, after skip=5mm]
\small
I will provide a passage of text, and you need to generate three different questions based on the content of this passage. Each question should be answerable using the information provided in the passage. Additionally, please provide an appropriate answer for each question derived from the passage.\\
You need to generate the question and answer in the following format:\\
{[}\\
\hspace*{1.5em}\{\\
\hspace*{3em}\textquotedbl{}question\textquotedbl{}: \textquotedbl{}What is the capital of France?\textquotedbl{},\\
\hspace*{3em}\textquotedbl{}answer\textquotedbl{}: \textquotedbl{}Paris\textquotedbl{}\\
\hspace*{3em}\textquotedbl{}full\_answer\textquotedbl{}: \textquotedbl{}The capital of France is Paris.\textquotedbl{}\\
\hspace*{1.5em}\},\\
{]}\\
This list should have at least three elements. You only need to output this list in the above format.\\
Passage:\\
\{passage\}
\end{tcolorbox}

\subsection{Prompt Instantiation}

For each example, the placeholders \verb|{context}|, \verb|{question}| are replaced with the actual content. Specially for multiple-choice datasets, placeholders \verb|{c0}|–\verb|{c4}| are replaced
with the corresponding candidate options. For example:
\begin{itemize}
    \item In the \verb|{context}| slot, insert the truncated, concatenated passages relevant to the question.
    \item In the \verb|{question}| slot, insert the input question from the dataset.
    \item In the  \verb|{c0}|–\verb|{c4}| slot, insert the choices relevant to the question.

\end{itemize}
This unified prompt design ensures that all methods are evaluated under consistent formatting conditions, enabling a fair and controlled comparison across baselines and proposed approaches.

\end{document}